\documentclass[11pt]{article}
\usepackage[a4paper]{geometry}
\usepackage{solar}
\usepackage{tabularx}
\usepackage{booktabs}
\usepackage{multirow}
\usepackage{graphicx}
\usepackage{hyperref}
\usepackage{amsmath}
\usepackage{amssymb}
\usepackage{listings}
\usepackage{color}

\usepackage{kotex}
\setmainhangulfont[Path=./]{NotoSansKR-Regular.ttf}
\setsanshangulfont[Path=./]{NotoSansKR-Regular.ttf}

\usetikzlibrary{arrows.meta,fit,positioning,calc,backgrounds}

\usepackage{xspace}
\newcommand{\myxspace}{\xspace}
\newcommand{\modelname}{{\sffamily Solar Open 2}\myxspace}

\graphicspath{{imgs/}}

\usepackage[most]{tcolorbox}
\usepackage{caption}
\usepackage{enumitem}

\title{Solar Open 2 Technical Report}

\author{
  Upstage Solar Team}
\date{Jul 22, 2026}
\website{https://upstage.ai}

\begin{document}
\sloppy
\maketitle

\vspace{-1.2cm}
\begin{abstract}

We present \modelname, a 250B-A15B Mixture-of-Experts language model built for long-horizon agentic tasks, scaled up from Solar Open 1 (Solar Open 100B).
To hold entire agent trajectories in a single context, \modelname reaches a 1M-token window through a hybrid attention stack that interleaves one softmax layer among every three linear-attention layers, using no positional encoding and a gated delta rule extended to negative eigenvalues.
To train at this scale under a fixed compute budget, we make training efficient in two ways: a stronger starting point, and higher-value data.
For the starting point, we initialize \modelname from Solar Open 1, transferring the 5.69B-parameter shared skeleton that survives the architectural change and learning everything else through full pre-training.
For the data, we curate for value per token: quality- and rarity-aware data curation and mixture-ratio optimization refine a 20T pool into a 10T mixture that, at equal token budget, outperforms the Solar Open 1 recipe.  
To build its agent skills, we train twelve domain specialists across purpose-built scenarios, then consolidate them into a single model by Multi-teacher On-Policy Distillation (MOPD).
Against comparably sized open-weight models on English benchmarks, \modelname leads on MMLU-Pro, LiveCodeBench, and the APEX-Agents agentic suite, and stays competitive with the strongest (DeepSeek-V4-Flash and MiMo-V2.5) elsewhere. On Korean benchmarks, \modelname records the highest average of any model compared, including fast-tier closed APIs, and on Ko-GDPval, an in-house Korean officework-agent benchmark, it is competitive with DeepSeek-V4-Pro (1.6T) at less than a sixth of its size.

\end{abstract}

\vspace{-1cm}

\begin{figure}[h!]
\centering
\includegraphics[width=\textwidth]{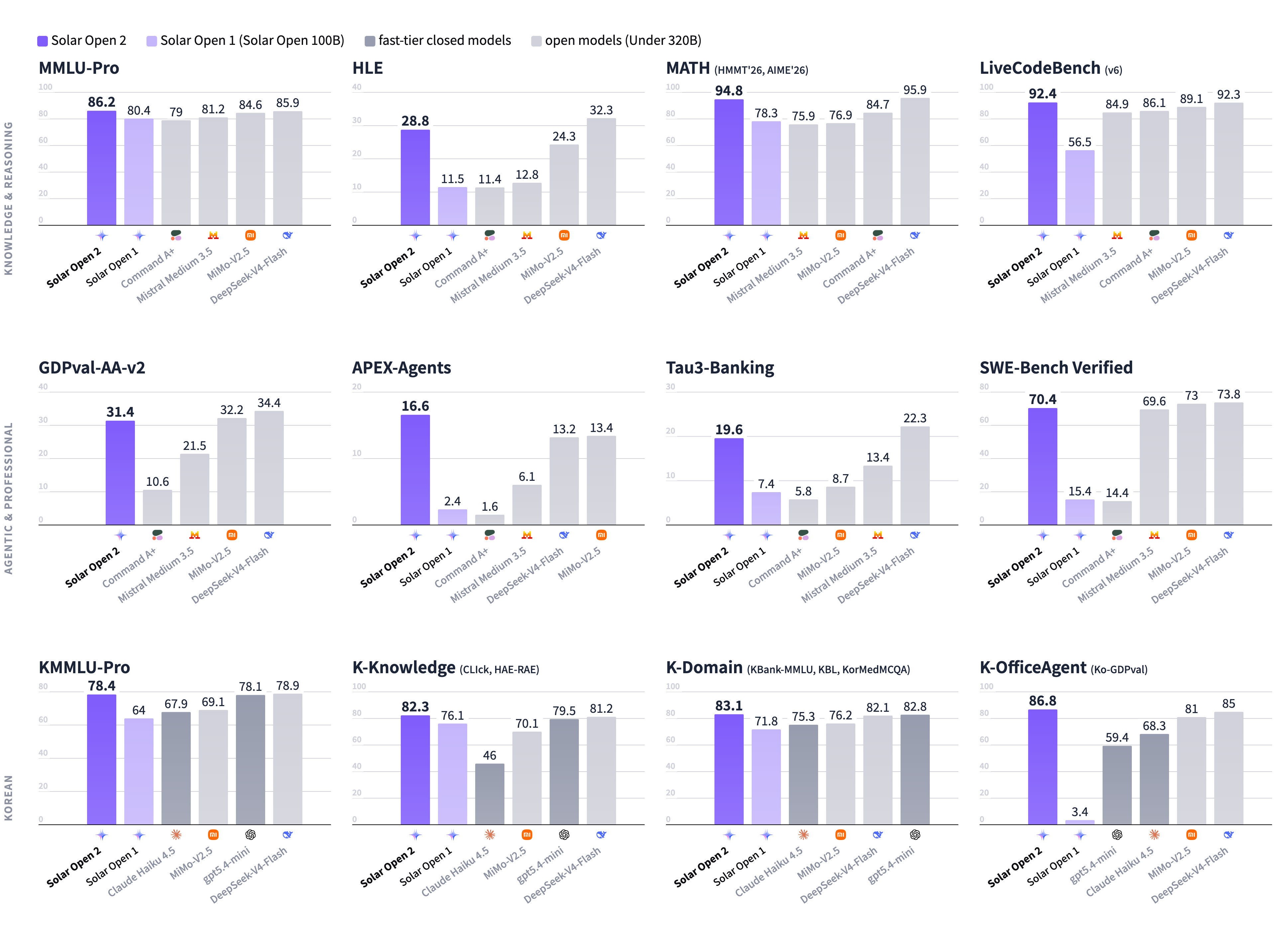}
\vspace{-1.0cm}
\caption{Benchmark comparison across three capability groups: knowledge and reasoning (top), agentic (middle), and Korean (bottom). 
}
\label{fig:benchmark_teaser}
\end{figure}
\section{Introduction}\label{sec:introduction}

Frontier language models have become a matter of national capability. For Korea, this makes a sovereign AI a strategic priority: an openly available, nationally grounded frontier model with strong native-language ability.
The field's shift toward long-horizon agents that plan and act over extended contexts raises this bar further, demanding both stronger reasoning and far longer context.
Solar Open 1 (102B-A12B) took a first step toward such a model, using a Korean-efficient tokenizer to reach strong Korean performance at modest scale~\citep{park2026solaropentechnicalreport}. In long-horizon agent settings, however, it fell short on two counts: limited long-context capability and thin coverage of agent scenarios.
 
We present \textbf{\modelname}, a 250B-A15B Mixture-of-Experts model that closes these gaps and advances Solar Open 1 along four axes.
First, it \emph{scales up} from 102B to 250B total parameters, raising the model's capacity for knowledge and reasoning.
Second, it \emph{preserves Korean token efficiency}, inheriting the Solar Open 1 tokenizer, for which global-model tokenizers spend 1.2--1.9$\times$ more tokens on the same Korean text.
Third, it \emph{adopts linear attention} within a hybrid attention stack (Section~\ref{sec:moe}), extending usable context from Solar Open 1's 128K to 1M tokens.
Fourth, it \emph{trains on purpose-built long-horizon agent scenarios} spanning conversational tool use, coding, and officework (Section~\ref{sec:post-training}), extending the sovereign model from language to agency.
Two further contributions keep training at this scale efficient: a selective weight transfer that initializes only about 2.3\% of \modelname's parameters from Solar Open 1, the portion that survives the architectural change (Section~\ref{sec:warmstart}), and data curation that maximizes value per token over a globally deduplicated corpus (Section~\ref{sec:pretrain-data}).
 
The result is a model strong in both languages and in agentic work. On Korean knowledge, reasoning, and agent benchmarks, \modelname records the highest average among comparably sized open-weight models and even fast-tier closed APIs. On Ko-GDPval, a Korean officework-agent benchmark, it essentially matches DeepSeek-V4-Pro (1.6T) at less than a sixth of its size. On English benchmarks, it leads comparably sized open-weight models on MMLU-Pro, LiveCodeBench, and the agentic APEX-Agents suite, and stays competitive with the strongest of them, DeepSeek-V4-Flash and MiMo-V2.5, elsewhere (Figure~\ref{fig:benchmark_teaser}; Section~\ref{sec:evaluation}).

\section{Model Architecture}\label{sec:architecture}

\modelname is a sparse Mixture-of-Experts (MoE) Transformer with 250B total parameters and 15B active parameters per token. 
It keeps the Solar Open 1 backbone unchanged: 48 layers, hidden size 4,096, head dimension 128, 64 query and 8 key-value heads, a 196,608-token vocabulary, a single shared expert with no dense layers, and pre-norm residuals. 
On this backbone it introduces three structural changes: a hybrid attention stack that interleaves linear and softmax attention, no positional encoding, and an expert pool enlarged from 128 to 320 routed experts.
The initialization of this architecture is itself a contribution, a selective weight transfer from Solar Open 1 described in Section~\ref{sec:warmstart}.

\subsection{\modelname Tokenizer}\label{sec:tokenizer}

\begin{figure}[htbp]
\centering
\includegraphics[width=\linewidth]{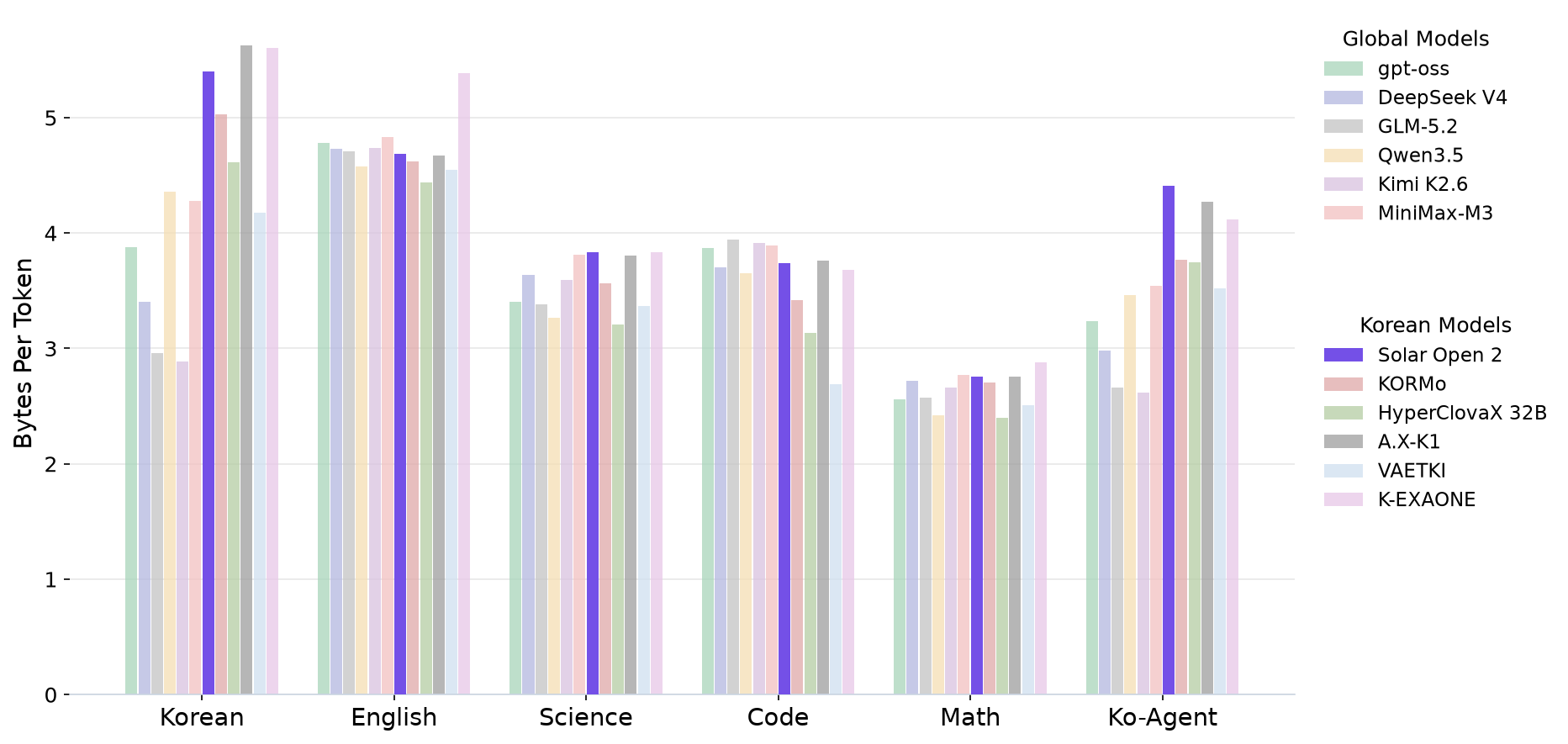}\\[2pt]
\caption{Tokenizer efficiency in bytes per token (higher is better: fewer tokens for the same text). \modelname inherits the Solar Open 1 tokenizer unchanged. In the Ko-Agent group (Ko-GDPval task prompts), it ranks first among the 12 tokenizers compared, about 24\% above the best global model (MiniMax-M3, 3.54).}
\label{fig:tokenizer_efficiency}
\end{figure}

\modelname inherits the Solar Open 1 tokenizer without modification. 
It is a custom byte-level BPE tokenizer with a 196,608-token vocabulary, trained on a corpus that oversamples Korean and the target domains, with digit splitting and whitespace preservation for arithmetic and code fidelity.
An identical vocabulary across the two generations is also a precondition for transferring the embedding and output-layer weights (Section~\ref{sec:warmstart}).

The tokenizer's strength is most pronounced in Korean, and especially in agent trajectories. 
Measured on the task prompts of Ko-GDPval, an in-house Korean officework-agent benchmark, it reaches 4.41 bytes per token, first among the 12 tokenizers compared (Figure~\ref{fig:tokenizer_efficiency}).
This is roughly 24\% more efficient than the best global model (MiniMax-M3, 3.54) and ahead of the Korean sovereign models A.X-K1 (4.27) and K-EXAONE (4.12).
The margin holds on general Korean text, where it reaches 5.40 bytes per token against 4.36 for the best global model (Qwen3.5).
In long-horizon agent scenarios, where the working context accumulates over many turns, expressing the same content in fewer tokens translates directly into lower inference cost and a longer effective context.

\subsection{\modelname Architecture}\label{sec:moe}

The central objective of the \modelname architecture is a usable context window beyond 1M tokens, served at approximately one quarter of the memory and computation of an all-softmax stack. Softmax attention grows its KV cache linearly and its computation quadratically with sequence length, becoming a bottleneck precisely in the long-horizon agent scenarios we target.

\begin{figure}[t!]
\centering
\includegraphics[width=\linewidth]{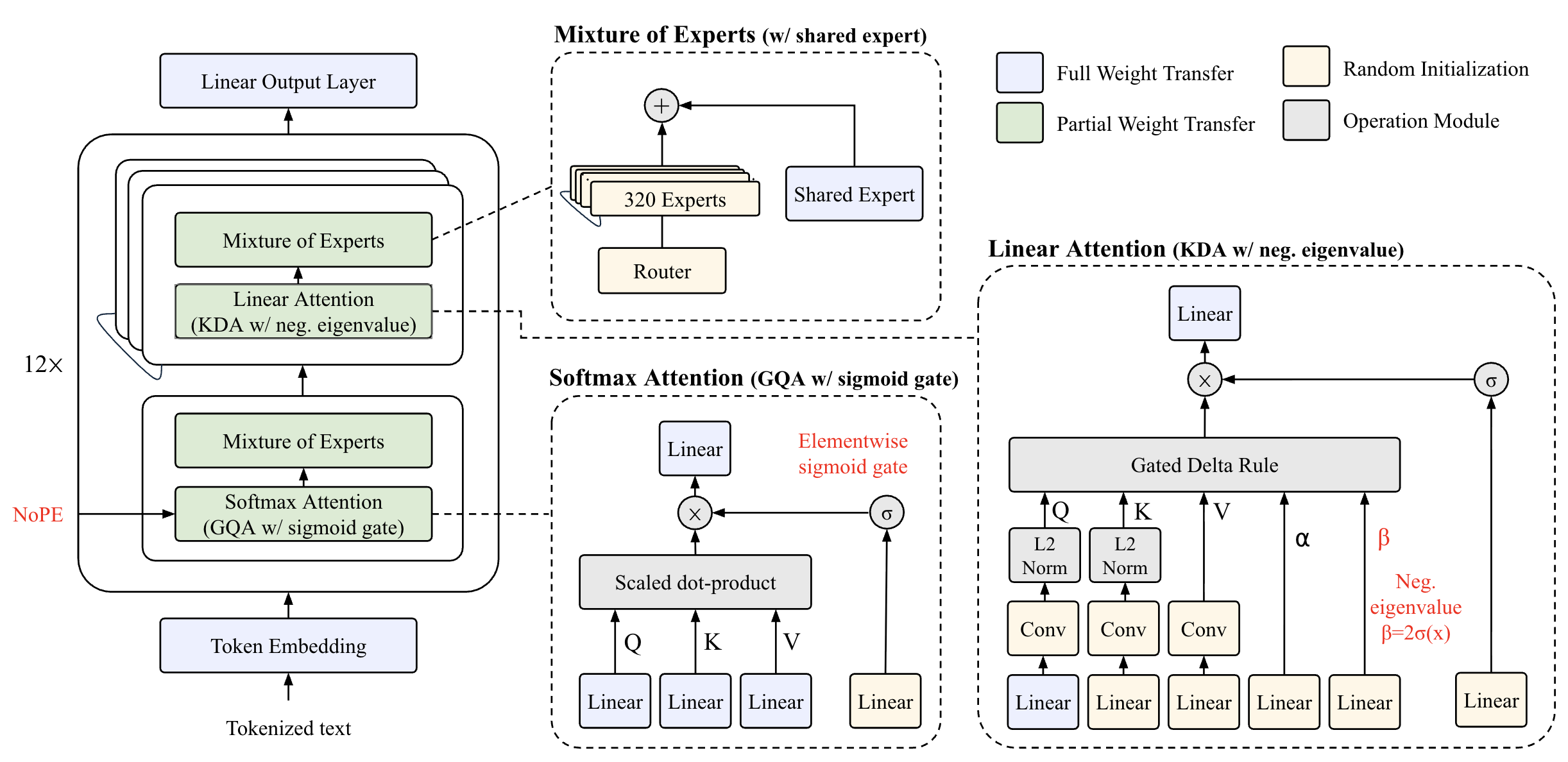}
\caption{\modelname architecture. The 48-layer stack (left) repeats a four-layer pattern twelve times: one softmax-attention layer followed by three linear-attention layers, each paired with an MoE block, and no positional encoding anywhere in the model. Insets detail the MoE block with 320 routed experts and one shared expert (top), the softmax-attention layer, GQA with an elementwise sigmoid output gate (bottom), and the linear-attention layer, KDA with the negative-eigenvalue extension $\beta = 2\sigma(\cdot) \in (0,2)$ (right). Colors mark the selective weight transfer (Section~\ref{sec:warmstart}): blue modules are transferred from Solar Open 1 in full, hatched modules in part, and yellow modules are randomly initialized.}

\label{fig:architecture_diagram}
\end{figure}

We address this bottleneck with a hybrid architecture that combines linear and softmax attention, following recent designs such as Kimi Linear~\citep{kimiteam2025kimilinear}, Qwen3.5~\citep{qwen2026qwen35}, and Nemotron~3~\citep{nvidia2025nemotron3}.
\modelname interleaves one softmax-attention layer with three linear-attention layers in every block of four and repeats this pattern twelve times, so that 12 of the 48 layers (25\%) are softmax and 36 (75\%) are linear (Figure~\ref{fig:architecture_diagram}). 
Within each block the softmax layer comes first (S-L-L-L), unlike the linear-first ordering (L-L-L-S) adopted by Kimi Linear and Qwen3.5.
It adopts GQA~\citep{ainslie2023gqa} for the softmax layers and KDA (Kimi Delta Attention)~\citep{yang2025gateddelta, kimiteam2025kimilinear} for the linear layers, with three extensions: no positional encoding (NoPE), a sigmoid output gate, and negative eigenvalues.
The linear layers carry the bulk of the sequence modeling: they accumulate context into a fixed-size recurrent state, keeping the KV cache constant in sequence length and the computation linear, while the softmax layers preserve the exact global recall that pure linear attention lacks.

\paragraph{No positional encoding (NoPE).} In softmax attention, positional encoding injects position information into the queries and keys so that attention can depend on relative distance.
Explicit schemes such as RoPE, however, tie the model to the length distribution seen during training, requiring data that spans the target range and capping generalization beyond the trained length~\citep{kazemnejad2023nope}.
\modelname removes positional encoding entirely: the softmax layers learn no explicit position signal, and relative position is instead carried solely by the linear-attention layers, which encode token order intrinsically through their sequential state.
This removes both the RoPE extrapolation limit and the softmax position-learning bottleneck. In principle, \modelname can therefore support an unbounded context length regardless of the length distribution of its training data.

\paragraph{Sigmoid output gate.} Following~\citet{qiu2025gated}, we add an elementwise sigmoid gate on the output of scaled dot-product attention in the softmax layers. Multiplying each head's output by this query-dependent gate (i) introduces non-linearity into the otherwise low-rank softmax-attention mapping, (ii) makes the output query-dependently sparse, and (iii) suppresses the ``attention sink'', the pathological collapse of attention onto a few early tokens. Empirically this gating improves training stability (tolerating larger learning rates) and strengthens long-context extrapolation~\citep{qiu2025gated}, both of which directly serve the 1M-token objective.

\paragraph{Negative eigenvalues.} Standard linear-attention cores, including the KDA used in Kimi Linear (\texttt{allow\_neg\_eigval=False}), restrict the eigenvalues of their state-transition matrix to [0,1], so the recurrent state can only decay or persist, never flip sign or actively erase.
Such cores provably cannot solve state-tracking problems such as parity or modular counting~\citep{grazzi2025unlocking}, and in practice, information written incorrectly into the fixed-size state persists and compounds along the sequence.
\modelname instead sets \texttt{allow\_neg\_eigval=True}: widening the write strength to $\beta = 2\sigma(\cdot) \in (0,2)$ (applied identically to the delta rule's erase term $\beta k k^\top S$ and write term $\beta k v^\top$) expands the eigenvalue range to $[-1,1]$.
Negative eigenvalues let the state flip sign and self-correct, restoring the capacity for genuine state tracking~\citep{grazzi2025unlocking}. 
This extension matters most in our linear-heavy, NoPE design: with no positional signal in the softmax layers, the linear state is the sole carrier of token order and long-range information, integrating over horizons of a million tokens. The ability to erase and invert, rather than merely decay, is what keeps that single long-lived state from drifting as errors accumulate.

\begin{table}[t!]
\small
\tabcolsep=5pt
\centering
\caption{Architecture specifications of Solar Open 1 (102B) and \modelname (250B). \modelname keeps the Solar Open 1 backbone dimensions while changing the attention stack, positional encoding, expert pool, context length, and initialization.}
\label{tab:architecture}
\resizebox{\textwidth}{!}{%
\begin{tabular}{l|ll}
\toprule
\textbf{Hyperparameter} & \textbf{Solar Open 1 (102B)} & \textbf{\modelname (250B)} \\
\midrule
Total Parameters & 102B & \textbf{250B} \\
Active Parameters / Token & 12B & \textbf{15B} \\
Context Length & 131,072 (128K) & \textbf{1,048,576 (1M)} \\
Vocabulary Size & 196,608 & 196,608 \\
\midrule
Num Layers & 48 & 48 \\
Hidden Size & 4,096 & 4,096 \\
Head Dimension & 128 & 128 \\
Attention Heads (Q / KV) & 64 / 8 & \textbf{64 / 8\,\textsuperscript{a}} \\
Attention & Softmax & \textbf{Hybrid: [Softmax $\times$1, Linear $\times$3] $\times$12\,\textsuperscript{b}} \\
Positional Embedding & RoPE ($\theta = 10^6$) & \textbf{NoPE} \\
\midrule
Num Experts (Routed) & 128 & \textbf{320} \\
Num Shared Experts & 1 & 1 \\
Experts per Token (Top-$k$) & 8 & 8 \\
MoE Intermediate Size & 1,280 & 1,280 \\
Activation & SiLU & SiLU \\
\midrule
Weight Init & from-scratch & \textbf{Selective transfer from Solar Open 1 (2.27\%)\,\textsuperscript{c}}  \\
\bottomrule
\end{tabular}%
}
\vspace{2pt}
\raggedright
{\footnotesize \textsuperscript{a} KV heads apply only to the softmax layers; the linear layers use 64 heads and keep no KV cache.} \\
{\footnotesize \textsuperscript{b} The softmax layers use GQA with a sigmoid output gate; the linear layers use an extension of KDA (see \S\ref{sec:moe}).} \\
{\footnotesize \textsuperscript{c} 5.69B weights (2.3\% of \modelname's 250B; 5.6\% of Solar Open 1's 102B) are initialized from Solar Open 1; the rest are random (see \S\ref{sec:warmstart}).}
\end{table}

\vspace{10pt}

Table~\ref{tab:architecture} summarizes architecture specifications of how \modelname differs from Solar Open 1. \modelname holds roughly two and a half times the parameters, with a different attention stack and more experts, yet the layer count and hidden dimension that form the shared backbone are kept identical. This is deliberate that the selective weight transfer reuses Solar Open 1's parameters exactly where the two models coincide (Section~\ref{sec:warmstart}).

\begin{figure}[t!]
\centering
\includegraphics[width=\linewidth]{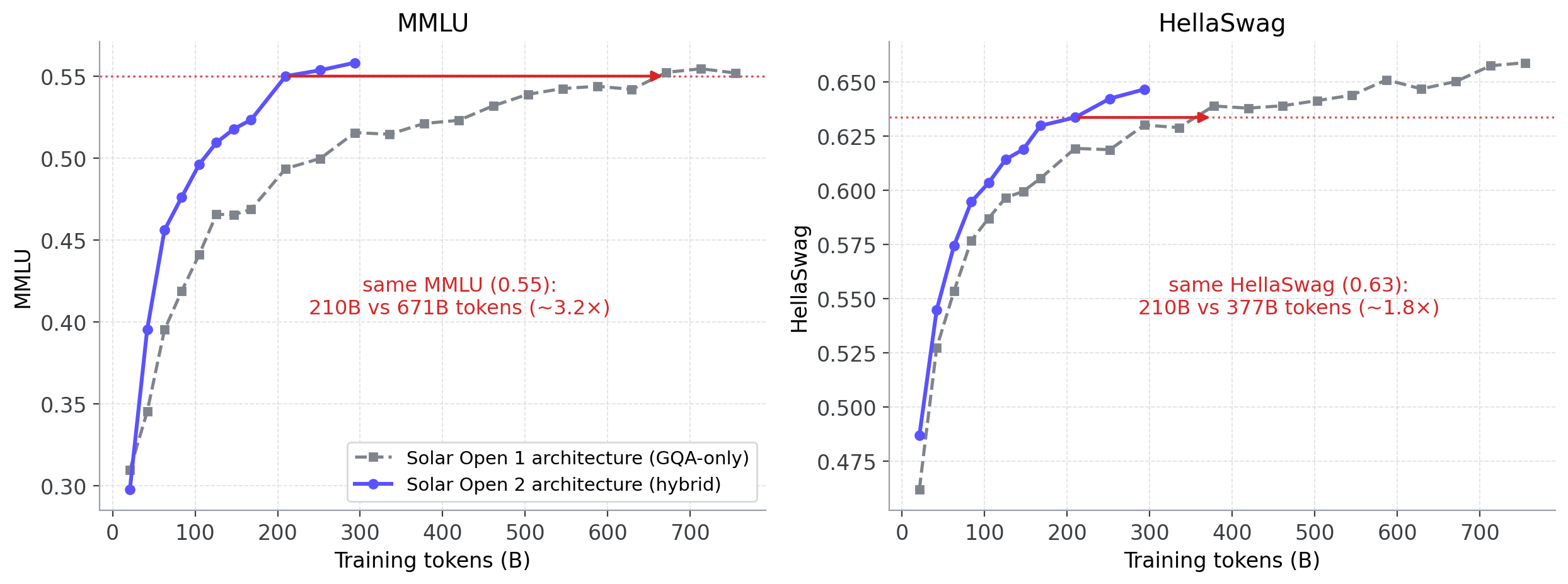}
\caption{Architecture ablation on a 10B-A1B proxy trained from scratch; only the architecture differs between the two runs. The hybrid \modelname stack reaches the MMLU 0.55 level with 210B training tokens, where the GQA-only Solar Open 1 stack needs 671B (3.2$\times$), and the HellaSwag 0.63 level with 210B versus 377B (1.8$\times$).}
\label{fig:arch_ablation}
\end{figure}

We ablated each of these changes on a 10B-A1B proxy before committing to the final design.
Relative to the all-softmax GQA architecture of Solar Open 1, the largest single gain comes from adopting linear attention together with NoPE: reaching a given validation loss consumes about 82.6\% of the baseline's tokens.
The remaining choices contribute smaller but consistent gains: the sigmoid output gate reduces the tokens needed to reach a given loss by roughly 3\%, the softmax-first ordering by about 1.5\%, and negative eigenvalues by about 1\%.
These loss-level margins appear modest, but small differences in loss translate nonlinearly into large differences on downstream benchmarks. Comparing the finalized \modelname architecture against the Solar Open 1 architecture at the same proxy scale, reaching the MMLU 0.55 level requires 210B training tokens where the baseline needs 671B, and the HellaSwag 0.63 level 210B against 377B (Figure~\ref{fig:arch_ablation}). 
The \modelname architecture is thus efficient on both fronts: attention cost at inference, and the training tokens required to reach a given capability.

\section{Pre-training}\label{sec:pre-training}

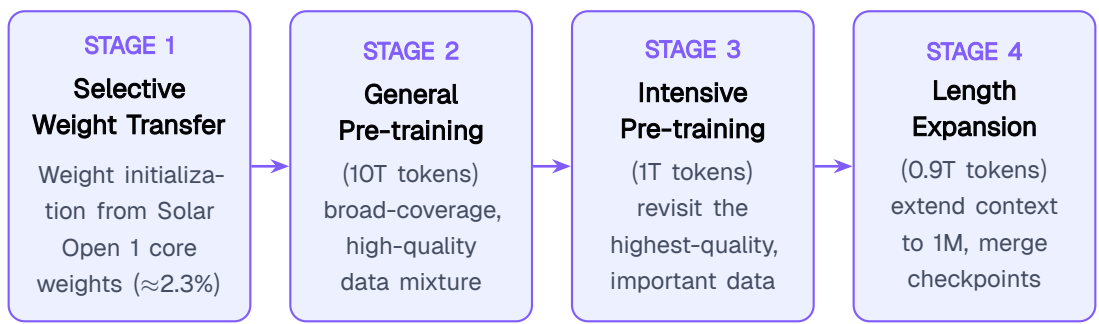
\begin{figure}[htbp]
\centering
\begin{tikzpicture}[
  font=\sffamily,
  stage/.style={
    draw=SolarViolet, line width=0.8pt, rounded corners=6pt,
    fill=JacksonBlue50, text width=2.7cm, align=center,
    inner sep=7pt, minimum height=4.1cm
  },
  arrow/.style={-{Stealth[length=2.5mm]}, line width=1pt, SolarViolet},
  node distance=0.5cm
]
\node[stage] (s1) {%
  {\footnotesize\color{SolarViolet}\bfseries STAGE 1}\\[3pt]
  {\small\bfseries Selective Weight Transfer}\\[5pt]
  {\footnotesize\color{SolarGray} Weight initialization from Solar Open 1 core weights ($\approx$2.3\%)}};
\node[stage, right=of s1] (s2) {%
  {\footnotesize\color{SolarViolet}\bfseries STAGE 2}\\[3pt]
  {\small\bfseries General Pre-training}\\[5pt]
  {\footnotesize\color{SolarGray} (10T tokens)\\ broad-coverage, high-quality data mixture}};
\node[stage, right=of s2] (s3) {%
  {\footnotesize\color{SolarViolet}\bfseries STAGE 3}\\[3pt]
  {\small\bfseries Intensive Pre-training}\\[5pt]
  {\footnotesize\color{SolarGray} (1T tokens)\\ revisit the highest-quality, important data}};
\node[stage, right=of s3] (s4) {%
  {\footnotesize\color{SolarViolet}\bfseries STAGE 4}\\[3pt]
  {\small\bfseries Length\\ Expansion}\\[5pt]
  {\footnotesize\color{SolarGray} (0.9T tokens)\\ extend context to 1M, merge checkpoints}};
\draw[arrow] (s1) -- (s2);
\draw[arrow] (s2) -- (s3);
\draw[arrow] (s3) -- (s4);
\end{tikzpicture}
\caption{Pre-training procedure of \modelname. Stage 1 initializes the model by selective weight transfer from Solar Open 1; Stages 2--4 then train on about 12T tokens in total: General Pre-training (10T), Intensive Pre-training (1T), and Length Expansion (0.9T) with checkpoint merging.}
\label{fig:pretrain_procedure}
\end{figure}

Four stages take \modelname from weight initialization to a 1M-context pre-trained model (Figure~\ref{fig:pretrain_procedure}): (1) Selective Weight Transfer, (2) General Pre-training, (3) Intensive Pre-training, and (4) Length Expansion.
Selective Weight Transfer initializes \modelname by transplanting the compatible core of Solar Open 1's weights (Section~\ref{sec:warmstart}).
General Pre-training is the main stage, training on 10T tokens of a broad-coverage, high-quality data mixture (Section~\ref{sec:pretrain-data}).
Intensive Pre-training then re-filters the same corpus at a substantially higher quality threshold and trains for another 1T tokens on the retained subset.
Length Expansion closes pre-training with 0.9T tokens of continued training that extend the context window to 1M and inject reasoning and agent data. Merging four intermediate checkpoints then produces the final pre-trained model.

\subsection{Selective Weight Transfer}\label{sec:warmstart}

Training a several-hundred-billion-parameter model from scratch is costly, so \modelname is instead initialized from the weights of its previous generation.
\modelname transfers weights directly and selectively from Solar Open 1, without distillation, and absorbs the architectural transition, including the expansion from 128 to 320 experts, through large-scale full pre-training.
Prior model-reuse work grows the \emph{same} architecture to a larger size~\citep{chen2022bert2bert, wang2023ligo, samragh2024hypercloning}, upcycles a dense checkpoint into an MoE~\citep{komatsuzaki2023upcycling, nakamura2025dropupcycling}, or converts softmax attention into a linear or recurrent structure through distillation or alignment losses with only small-scale uptraining~\citep{wang2024mambollama, bick2024mohawk, chattopadhyay2026priming, mercat2024supra, zhang2024lolcats, lan2025liger, sse2025sparse}. To our knowledge, every recent large hybrid model has been pre-trained from scratch~\citep{kimiteam2025kimilinear, qwen2025qwen3next, minimax2025lightning}.
None of these approaches covers that combination. We call it \emph{selective weight transfer}: in effect, a partial warm start carried across an architectural change.

The selection principle is simple: transfer the weights whose representations are shared between the two models, and randomly initialize everything whose shape or computation changed (Figure~\ref{fig:architecture_diagram}, Table~\ref{tab:warmstart_transfer}). 
Only the compatible portion is reused: about 2.3\% of parameters, or 5.69B.
The token embedding and output layer transfer because the vocabulary is identical (196,608), letting learned token semantics carry over; the normalization layers transfer for training stability; and the softmax-attention $q/k/v/o$ projections transfer unchanged because \modelname keeps Solar Open 1's GQA geometry (64 query and 8 KV heads).
In the linear-attention layers, only the query and output projections are re-mapped from the GQA $q/o$ weights, whose shapes match (64 heads); the key and value projections cannot transfer, since GQA keeps 8 KV heads while the linear layers use 64 MHA-style heads, and the decay and write gates and short convolutions have no counterpart in Solar Open 1. The sigmoid output gate of the softmax layers is likewise new and randomly initialized.
Finally, the shared expert transfers: its shape is unchanged, and as an always-active general-purpose path it carries over cleanly. The 320 routed experts are instead randomly initialized, since expanding from 128 to 320 experts admits no one-to-one mapping, and replicating experts to fill the gap risks growing inter-expert similarity at the cost of specialization.
As Table~\ref{tab:warmstart_transfer} shows, the transferred 5.69B parameters are exactly the model's shared computational skeleton, and the randomly initialized remainder is almost entirely the routed experts (241.7B of 244.3B).

\begin{table}[t!]
\small
\centering
\caption{Selective weight transfer: per-module weight handling. The two architectures differ structurally, but their shared skeleton is kept aligned: within each module, the shape-compatible weights are transferred (\checkmark) and only the genuinely new or incompatible parts are randomly initialized ($\times$). Transferred components sum to 5.69B of the 250B total; the randomly initialized remainder is almost entirely the routed experts (241.7B).}
\label{tab:warmstart_transfer}
\begin{tabular}{@{}llrc@{}}
\toprule
\textbf{Component} & \textbf{Shape relation} & \textbf{Params.} & \textbf{Init.} \\
 & (Solar Open 1 $\rightarrow$ \modelname) & & \\
\midrule
Token embedding / output layer & identical & 1.61B & \checkmark\ transfer \\
Normalization layers & identical & $<$0.01B & \checkmark\ transfer \\
\addlinespace[3pt]
\multicolumn{4}{@{}l}{Softmax attention (12 layers)} \\
\quad $q/k/v/o$ projections & identical & 0.91B & \checkmark\ transfer \\
\quad sigmoid output gate & new & $<$0.01B & $\times$ random \\
\addlinespace[3pt]
\multicolumn{4}{@{}l}{Linear attention (36 layers)} \\
\quad $q/o$ projections & new; re-mapped from GQA $q/o$ & 2.42B & \checkmark\ transfer \\
\quad $k/v$ proj., decay/write gates, conv & new & $\approx$2.5B & $\times$ random \\
\addlinespace[3pt]
\multicolumn{4}{@{}l}{MoE FFN (48 layers)} \\
\quad shared expert & identical & 0.75B & \checkmark\ transfer \\
\quad routed experts ($\times$320) + router & $128 \rightarrow 320$, no 1:1 mapping & $\approx$241.7B & $\times$ random \\
\midrule
\textbf{Transferred total} & & \textbf{5.69B} & \\
\bottomrule
\end{tabular}
\end{table}

\begin{figure}[htbp]
\centering
\includegraphics[width=0.7\linewidth]{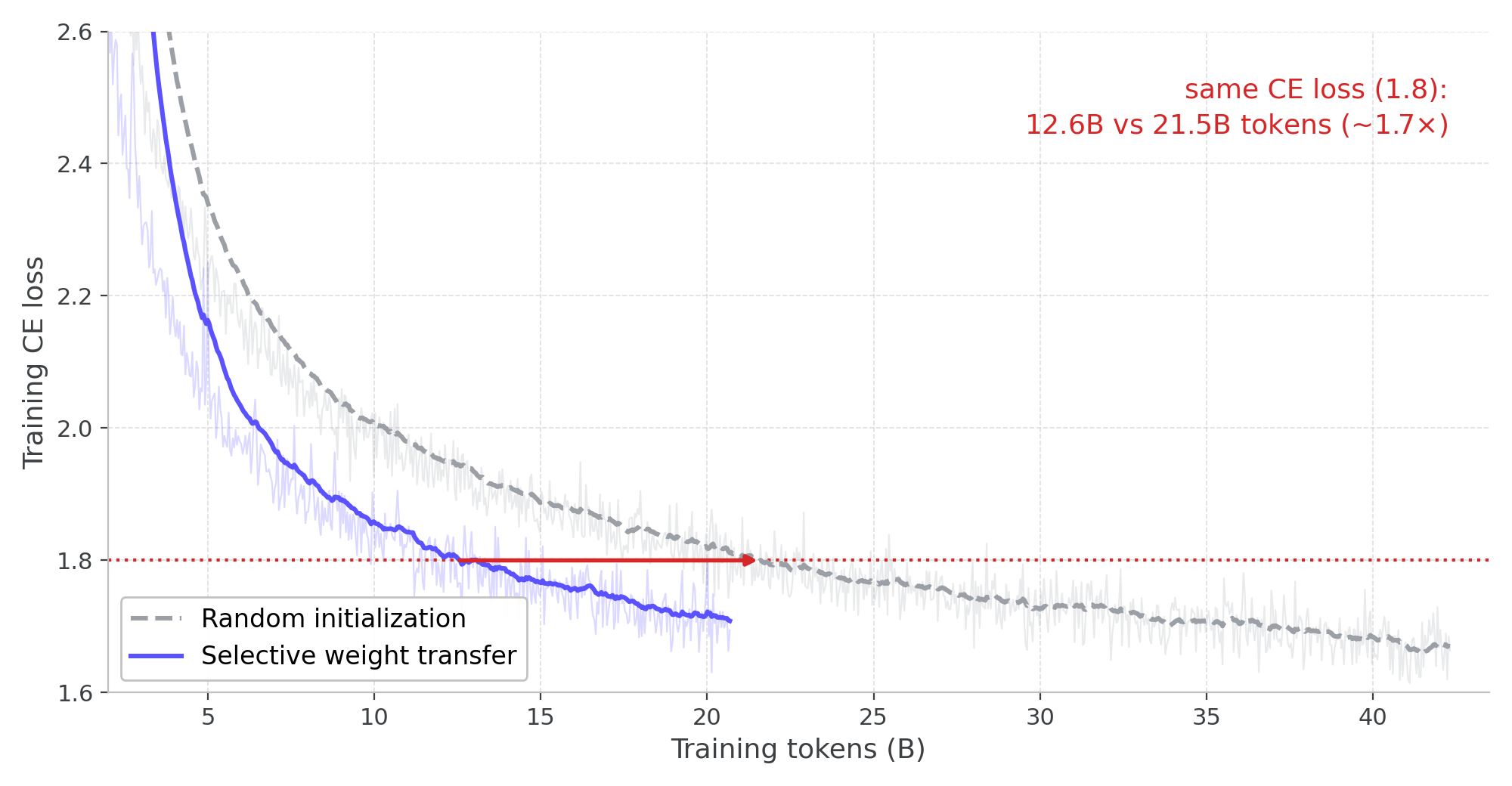}
\caption{Selective weight transfer vs.\ random initialization on a \modelname 200B-A15B proxy (controlled: only the initialization differs; same architecture, data, and optimizer), zoomed to the 1.6--2.6 loss band. Transferring 5.69B parameters from Solar Open 1 reaches a training cross-entropy of 1.8 in 12.6B tokens where random initialization needs 21.5B ($\sim$1.7$\times$). Curves are EMA-smoothed ($\alpha = 0.06$); the raw traces are shown faintly behind them.}
\label{fig:warmstart_ablation}
\end{figure}

\begin{figure}[htbp]
\centering
\includegraphics[width=0.92\linewidth]{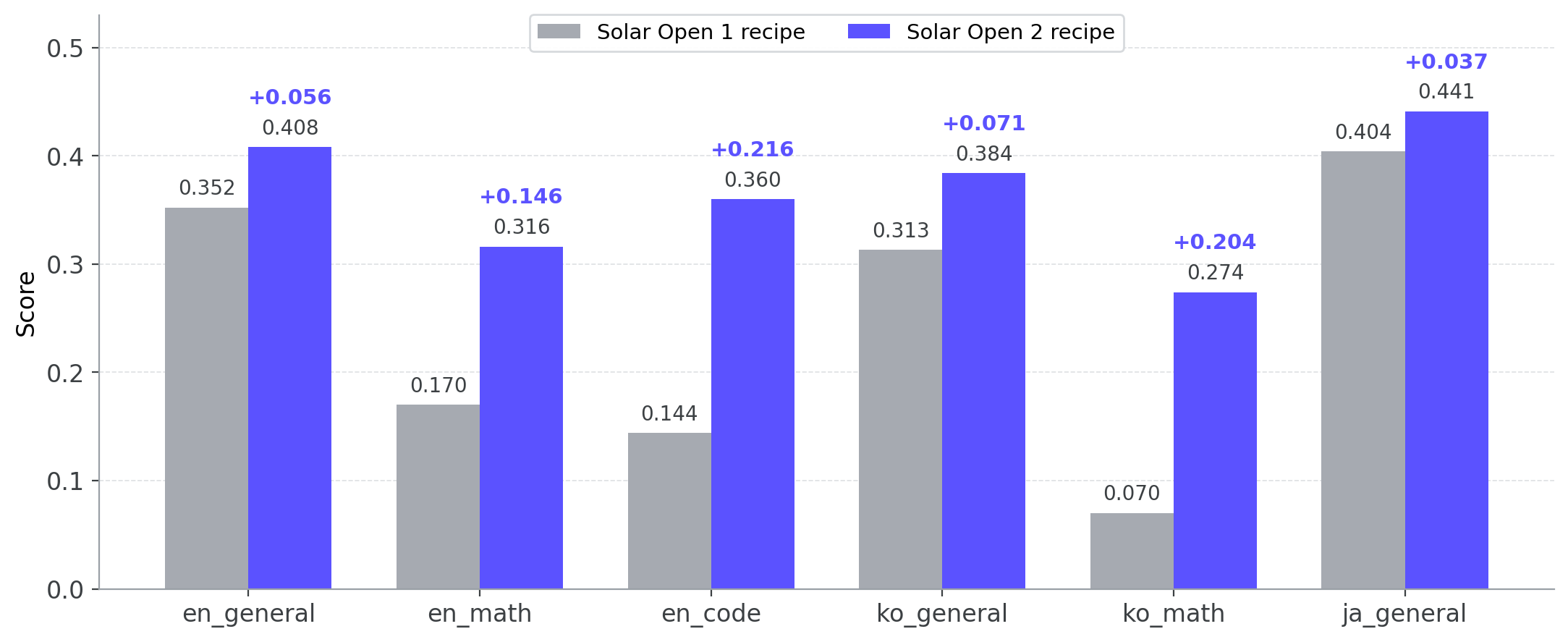}
\caption{Data-recipe ablation on a 10B MoE model: Solar Open 1 vs.\ \modelname recipes at an equal budget of 300B training tokens. The \modelname recipe leads on every benchmark group; the largest gains are en\_code (+0.216), ko\_math (+0.204), and en\_math (+0.146).}
\label{fig:perdomain}
\end{figure}

The benefit is clear on a 200B-A15B proxy in which only the initialization differs: one run starts from the 5.69B parameters carried over from Solar Open 1, the other from none. Reaching a training cross-entropy of 1.8 consumes 12.6B tokens with the transfer against 21.5B without --- about 1.7$\times$ fewer, or 58\% of the tokens (Figure~\ref{fig:warmstart_ablation}). The margin widens across the observed band, from roughly 1.15$\times$ at a loss of 2.6 to 1.7$\times$ at 1.8. 

\subsection{Pre-training Data Curation}\label{sec:pretrain-data}

For pre-training data, collecting large-scale raw data from diverse sources matters, but so does selecting what is meaningful and shaping it into a composition effective for training.
Where Solar Open 1 concentrated on the former, gathering data and training on it at scale, \modelname inherits the core of Solar Open 1's weights as its skeleton (Section~\ref{sec:warmstart}) and concentrates on the latter: an effective data composition built on top of that skeleton.

We construct a pipeline that further refines an initially cleaned pool of 20T tokens down to 10T.
Every dataset receives an in-house quality score, while its \emph{rarity} is tracked by source and type.
The pipeline removes both exact and semantic duplication~\citep{lee2022dedup, abbas2023semdedup}, and organizes the resulting pool under a classification scheme keyed by domain, language, and source.

The last lever is the mixture ratio, which we optimized through ablation studies.
At the top level, the resulting recipe sets (1) a real-to-synthetic ratio of 4:6, (2) a domain mix in which math and code each occupy at least 15\%, and (3) an English share above 80\%.
Figure~\ref{fig:perdomain} compares the Solar Open 1 and \modelname data recipes on a 10B MoE model at equal training budget (300B tokens): the \modelname recipe improves consistently on every benchmark group, with the largest gains on math and code.
This equal-budget comparison isolates value per token: with compute fixed, the gains come entirely from what each token teaches.

\subsection{Curriculum Learning}\label{sec:curriculum}

On top of the curated data, pre-training proceeds as a three-stage curriculum spanning Stages 2--4 of Figure~\ref{fig:pretrain_procedure}. 
Stage 2 (General Pre-training) trains on the 10T-token curated dataset.
Stage 3 (Intensive Pre-training) raises the quality threshold and trains again on the best 1T tokens of the Stage-2 data, a design inspired by repeated training on constrained high-quality data~\citep{muennighoff2023scaling} and by end-of-training domain upsampling~\citep{blakeney2024does}.
Stage 4 (Length Expansion) raises the quality threshold further to form its base corpus, augments it with long-document data including repo-level code, and extends training to a 1M context length.

\begin{figure}[htbp]
\centering
\includegraphics[width=\linewidth]{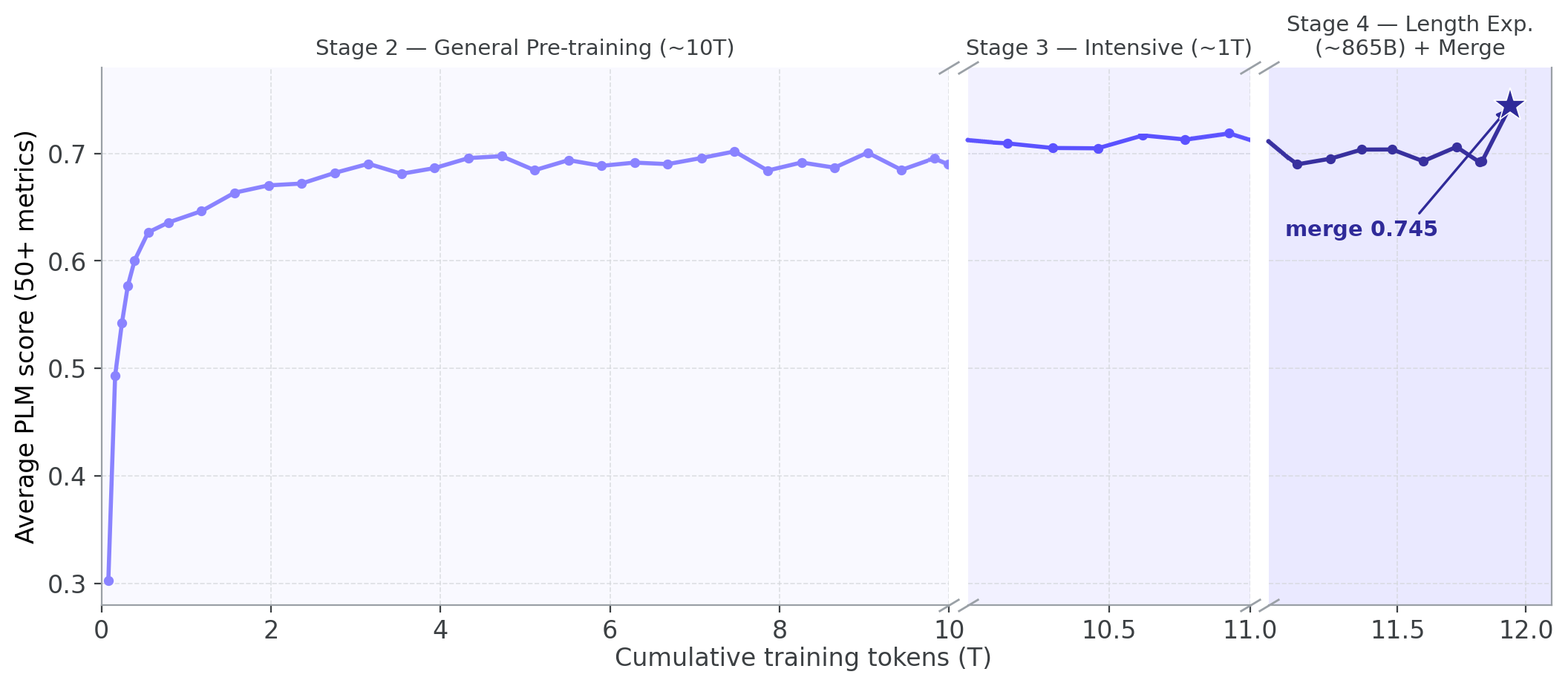}
\caption{Average performance over 50+ PLM evaluation metrics as a function of cumulative training tokens across the curriculum: Stage 2 (General Pre-training, 10T), Stage 3 (Intensive Pre-training, 1T), and Stage 4 (Length Expansion, 865B). Performance rises then stagnates during Stage 2, improves again in Stage 3, and slightly degrades during Stage 4. Star marks the final checkpoint merge, which yields the highest score (0.745).}
\label{fig:phase_performance}
\end{figure}

During Length Expansion, we observed an overall performance drop, which we attribute to the shift in data distribution.
To compensate, we selected four intermediate checkpoints and merged them, obtaining a pre-trained language model (PLM) stronger than any single checkpoint.
Figure~\ref{fig:phase_performance} traces the average of more than 50 PLM evaluation metrics as a function of training tokens across the stages: performance improves through Stage 2 but eventually stagnates, rises again as Stage 3 begins, and slightly degrades in Stage 4. 
Stars mark checkpoint merges; the final merge achieves the highest score (0.745).

\section{Post-training}\label{sec:post-training}

\begin{figure}[htbp]
\centering
\begin{tikzpicture}[
  font=\sffamily,
  stage/.style={
    draw=SolarViolet, line width=0.8pt, rounded corners=6pt,
    fill=JacksonBlue50, text width=2.7cm, align=center,
    inner sep=7pt, minimum height=4.1cm
  },
  arrow/.style={-{Stealth[length=2.5mm]}, line width=1pt, SolarViolet},
  node distance=0.5cm
]
\node[stage] (p1) {%
  {\footnotesize\color{SolarViolet}\bfseries STAGE 1}\\[3pt]
  {\small\bfseries Supervised\\ Fine-Tuning}\\[5pt]
  {\footnotesize\color{SolarGray} chat-template following plus target reasoning and agent tasks}};
\node[stage, right=of p1] (p2) {%
  {\footnotesize\color{SolarViolet}\bfseries STAGE 2}\\[3pt]
  {\small\bfseries Multi-domain\\ RL}\\[5pt]
  {\footnotesize\color{SolarGray} verifiable-reward RL on STEM-centered reasoning tasks}};
\node[stage, right=of p2] (p3) {%
  {\footnotesize\color{SolarViolet}\bfseries STAGE 3}\\[3pt]
  {\small\bfseries Specialist}\\[5pt]
  {\footnotesize\color{SolarGray} twelve teachers trained by SFT and domain-specific RL}};
\node[stage, right=of p3] (p4) {%
  {\footnotesize\color{SolarViolet}\bfseries STAGE 4}\\[3pt]
  {\small\bfseries Multi-teacher On-policy Distillation}\\[5pt]
  {\footnotesize\color{SolarGray} consolidate twelve teachers into one model}};
\draw[arrow] (p1) -- (p2);
\draw[arrow] (p2) -- (p3);
\draw[arrow] (p3) -- (p4);
\end{tikzpicture}
\caption{Post-training procedure of \modelname. SFT and Multi-domain RL build the generalist base; the Specialist stage then cultivates twelve domain teachers in parallel, and Multi-teacher On-policy Distillation (MOPD) consolidates them into the single released model.}
\label{fig:posttrain_procedure}
\end{figure}
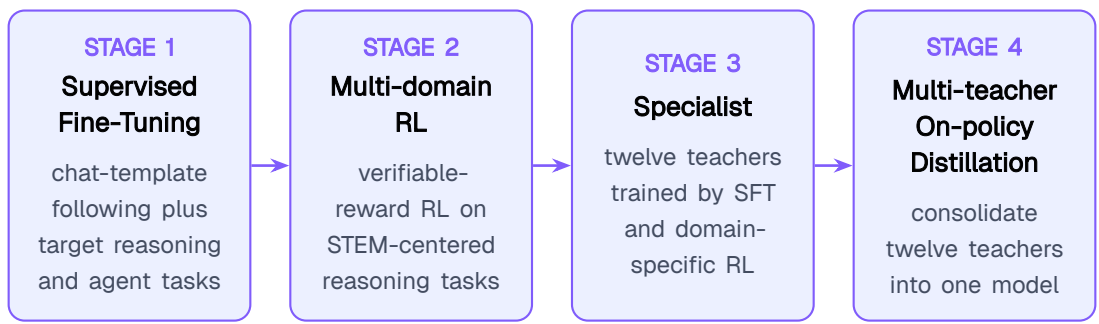

Post-training of \modelname proceeds in four stages (Figure~\ref{fig:posttrain_procedure}): (1) Supervised Fine-Tuning (SFT), (2) Multi-domain RL, (3) Specialist, and (4) Multi-teacher On-policy Distillation (MOPD). 
The SFT stage goes beyond basic chat-template following and targets the full range of reasoning and agent tasks \modelname is built to cover. 
Multi-domain RL then strengthens core reasoning with verifiable rewards over a broad set of STEM-centered tasks.
The Specialist stage trains twelve domain specialists, such as the coding and officework agents, through SFT and domain-specific RL to serve as teachers. The final stage, MOPD, folds these twelve teachers into a single model.
The first two stages follow established recipes, and their data compositions are outside the scope of this report.
The remainder of this section first describes the agent scenarios that supply the tasks, environments, and verification signals for these stages (Section~\ref{sec:agent-scenarios}), and then the two pieces of training infrastructure that execute the stages at 250B scale, fully asynchronous RL and MOPD (Section~\ref{sec:post-training-recipe}).

\subsection{Agent Scenarios}\label{sec:agent-scenarios}

The agent capabilities of \modelname are trained from purpose-built \emph{scenarios} rather than harvested corpora: for each target capability we construct the environment, synthesize tasks inside it, roll out trajectories, and admit an instance only when it passes execution- or evidence-based verification. Every pipeline is additionally decontaminated against the benchmarks used in Section~\ref{sec:evaluation}.
This section summarizes the three scenario families: general conversational agents, coding agents, and officework agents. For each, we describe which scenarios were prepared and how their correctness is guaranteed.

\subsubsection{General Conversational Agent}\label{sec:agent-conversational}

We define and develop the general conversational agent along two axes: tool execution over stateful Model Context Protocol (MCP) environments, and multi-turn, policy-bound conversation grounded in a live transactional domain. 
For the first axis, two bottlenecks dominate: (i) the environment itself, since a general pipeline needs either faithful simulation of stateful tools or the infrastructure to call real ones; and (ii) task instructions for state-mutating Create/Update/Delete (CUD) operations, for which almost no source material exists.
We compose a layered environment substrate spanning both faithful simulations of stateful tools and real callable services. The CUD gap is resolved by generating the \emph{verifier first}: the pipeline executes a mutation against the live environment, records its expected state effects, and synthesizes a read-back verifier, a sandboxed \texttt{pytest} suite over the environment's read tools, whose success defines the task's correctness. The instruction is then rewritten at a controlled obfuscation level, so the agent must discover the solution path from the environment rather than transcribe it.
Every task carries a graded process rubric and an executable end-state check, enforced through a three-pass self-consistency verification: hard trace rules, an LLM coherence judge, and the executable read-back.

For the second axis, we prepare a live transactional database governed by an explicit domain policy and run user--agent simulations over it. Tasks are instantiated from composite scenarios and pinned to entities that genuinely exist in the database; a user simulator reveals intent gradually across turns, withholding identity fields under failed-verification scenarios~\citep{cho2026useroriented}; and every tool call executes against the real database. Acceptance combines an LLM rubric judge with structural gates that encode the policy semantics a semantic judge tends to miss.

\subsubsection{Coding Agent}\label{sec:agent-coding}

The coding agent comprises three scenarios: \emph{software engineering} (SWE) over realistic repository environments, \emph{terminal} for real-world tasks beyond software engineering in live terminals, and \emph{artifact} for front-end generation.

The SWE scenario demands robustness to heterogeneous agent scaffolds and the ability to iteratively edit, execute, and debug code. We build it along three axes.
First, we employ diverse scaffolds (OpenHands, Claude Code, OpenCode, KiloCode, HermesAgent), since a model trained on a single scaffold overfits its tool schemas and interaction protocols. 
Second, we synthesize executable environments at scale. Pull requests mined from repositories across 18 programming languages are retained only when a complete task specification can be recovered: problem statement, gold patch, test patch, base commit, and environment requirements. An instance is then admitted only if its LLM-synthesized Docker environment is \emph{buildable} and \emph{fail-to-pass}, meaning the test script fails on the base commit and passes after the gold patch. Finally, the repository's entire Git history is stripped, so the ground-truth fix cannot be recovered from version control.
Third, trajectories rolled out under the diverse scaffolds in these synthesized environments pass a layered filter stack: malformed-tool-call filtering, a five-dimension LLM-judged quality rubric, and repository-level decontamination (strictly more conservative than instance-level filtering) against SWE-bench Verified, SWE-bench Multilingual, and SWE-bench Pro~\citep{jimenez2023swebench, deng2025swebenchpro}.

The \emph{terminal} scenario covers real-world work across more than ten domains, such as data science and scientific computing, executed in live terminal environments under multiple agent harnesses.
Trajectory synthesis deliberately includes encountering and recovering from runtime errors, and places particular weight on \emph{self-verification} behavior: the agent composes and runs code-level self-tests before declaring completion and, on failure, revises its work and re-tests until the suite passes.
Trajectories are filtered by rewriting or removing hallucinated content, by rubrics over task complexity and error patterns (task completion, repeated actions, acting without analysis), and by decontamination against Terminal-Bench~\citep{merrill2026terminalbench}.

The \emph{artifact} (front-end generation) scenario starts from category-wise task synthesis over six categories: website, game development, 3D design, data visualization, UI components, and SVG. A large LLM-generated seed pool is deduplicated and clustered by maximizing the Vendi Score~\citep{friedman2023vendi}. Each representative seed is expanded with a tech stack, UI/layout, and feature requirements, together with an expected-essential-feature list that is reused downstream as a functional rubric. LLM-as-a-judge screening then discards infeasible, stack-inconsistent, or under-specified tasks.
Trajectories are generated under category-specific guidance contexts (best practices, anti-clich\'e constraints, recommended library usage), which are simplified or removed at training time for context distillation. They then pass a two-stage rejection sampling: executability filtering (static lint plus Playwright headless-browser runtime checks), followed by VLM-as-a-judge screening of full-page screenshots for functional completeness against the essential-feature rubric and for visual completeness against category-specific checklists.

\subsubsection{Officework Agent}\label{sec:officework}

The officework agent is built to reliably carry out the tasks knowledge workers face in everyday office environments, which demands an unusually broad combination of capabilities: command of the domain knowledge and working conventions of many industries and occupations; the ability to interpret heterogeneous reference materials and produce work artifacts in office-native formats (xlsx/docx/pptx/pdf); and the workspace competence to gather the information a task actually depends on and keep every produced number consistent with the underlying data.
The central obstacle to training these capabilities is the acute scarcity of task data: real office work is proprietary, human demonstrations do not scale across hundreds of occupations, and existing agent corpora concentrate on web navigation, search, and software engineering.
To close this gap, we develop \textbf{OfficeVerse}, an in-house pipeline that automatically synthesizes office tasks grounded in real public data, together with a grading scheme that scores each task's success or failure. 
OfficeVerse organizes office work as a matrix of 11 industry domains $\times$ 12 task types, where each task type pairs a cognitive operation with a primary deliverable format (e.g.\ analysis$\rightarrow$xlsx, presentation$\rightarrow$pptx). Each cell holds reusable task specifications declaring the external data sources a task consumes, typed schemas for supplementary data, admissible data-sourcing policies, and output specifications.
The deliverable-format mix is calibrated against sector-level statistics measured from real professional work products. Every task is anchored in real, commercially usable public data, which a library of fetchers and converters collects and turns into the office-native reference files a professional would receive alongside an assignment.

Two complementary generation tracks feed the pipeline. 
\emph{Atomic occupational task synthesis} decomposes each occupation's daily work into single-deliverable assignments and instantiates them at effectively unbounded scale through a four-stage process: context generation, supplementary-data generation, prompt generation, and rubric generation. A deterministic validation gate follows every stage, so a task whose premise contradicts its own attachments is discarded at creation rather than repaired downstream. The resulting weighted rubrics combine mechanically checkable criteria with judge-graded open-ended ones, each cross-validated against ground truth extracted from the task's own materials.
\emph{World-grounded workspace synthesis} instead builds entire working environments: professional project trees seeded from real companies in document-intensive verticals, populated with heterogeneous materials, with the evidence deliberately dispersed across directories. It then synthesizes tasks whose prompts withhold file paths and other locators and whose answers span several files, so workspace retrieval and cross-file synthesis become the supervised skill. An environment-quality audit rejects any instance solvable by shortcut (a single obvious file, parametric recall, or degenerate reasoning).
On top of both tracks, task demand is controlled along two axes: (i) a \emph{difficulty} level re-derived from the constraints that actually survive into the final prompt; and (ii) an \emph{expertise} level realized as a curriculum that progressively strips scaffolding from over-scaffolded prompts and re-anchors the rubric toward the open expert judgment the leaner prompt implicitly demands.
Over these synthesized tasks, trajectories are rolled out by capable agent models under bounded turn and wall-clock budgets in track-matched harnesses: a sandboxed knowledge-worker machine with an office-document toolchain and a code-execution tool for atomic tasks, and the generated environment itself as the operating surface for workspace tasks. The trajectories are then serialized into a single canonical schema.
Post-processing deliberately separates process quality from outcome quality. Rule-based prefilters and an LLM judge gate data fabrication and misuse; deliverables are extracted from their binary formats, and each rubric criterion is routed to a rule-based checker or an LLM judge. The same outcome record that gates SFT admission is the source from which RL rewards are derived.

The full pipeline also covers Korean office work: a Korean occupational taxonomy, grounding in Korean public and industry data, enforcement of Korean business-writing conventions (native numeric units, standard domain terminology, Korean-style file naming), an English-reasoning/Korean-output language protocol with a language-identification filter, and CJK rendering-integrity checks down to detecting missing glyphs from the code that drew a chart. To our knowledge, the resulting corpus is the first Korean training set of deliverable-producing office tasks. A Korean officework task example is given in Section~\ref{sec:appendix:officework}.

\subsection{Training}\label{sec:post-training-recipe}

Two pieces of training infrastructure underpin post-training: (i) a fully asynchronous RL system (Section~\ref{sec:async-rl}), used throughout Multi-domain RL and the domain-specific RL of the Specialist stage; and (ii) the multi-teacher on-policy distillation setup (Section~\ref{sec:mopd}), used for the final consolidation.

\subsubsection{Fully Asynchronous Reinforcement Learning}\label{sec:async-rl}

Agentic RL rollouts exhibit a long-tailed length distribution: a small number of trajectories take far longer to complete than the rest. Under a synchronous regime, every optimizer step must wait for the slowest rollout in the batch, leaving the majority of devices idle. 
We therefore adopt a fully asynchronous design built on five components: (i) a disaggregated architecture with in-flight weight updates, (ii) staleness control via a fresh-token-fraction gateway, (iii) length- and staleness-aware batch sampling, (iv) direct double-sided importance sampling, and (v) environment-failure filtering and group-size repair.

\paragraph{Disaggregated asynchronous architecture.} The trainer and the actor (the rollout engine) are disaggregated onto separate GPU pools: the actor generates trajectories continuously, and whenever the rollout buffer holds enough completed trajectories, a batch is dispatched for a policy update. When the trainer publishes new parameters, in-flight generation is briefly suspended, the weights are broadcast, and the KV caches of all partially generated trajectories are recomputed under the new parameters before generation resumes. A single trajectory may therefore contain segments produced by several successive policy versions, a property the components below account for explicitly. The actor exchanges data with the trainer directly as token IDs with per-token metadata (policy version and rollout-time log-probabilities), eliminating detokenization and re-tokenization from the loop entirely, along with the silent training-signal corruption their boundary mismatches can cause.

\begin{figure}[htbp]
\centering
\includegraphics[width=0.78\linewidth]{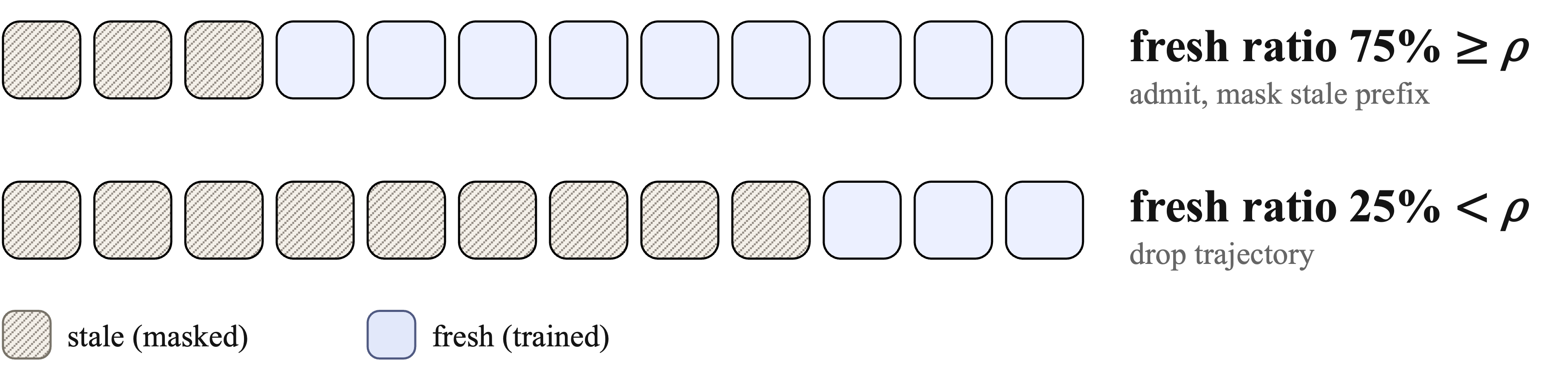}
\caption{The fresh-token-fraction gateway. Every token records the policy version that generated it, and a token is fresh if its staleness is at most $s_{\max}$. A trajectory whose fresh-token ratio reaches the threshold $\rho$ is admitted with its stale prefix masked from the loss (top); one whose ratio falls short is dropped entirely (bottom).}
\label{fig:fresh_gateway}
\end{figure}

\paragraph{Staleness control via a fresh-token-fraction gateway.} 
Asynchrony introduces off-policy drift, but dropping an entire trajectory as soon as any token exceeds a maximum staleness ($s_{\max}$) wastes the compute already invested in long rollouts.
We instead record, for every token, the version of the policy that generated it, and measure the token's staleness as the gap between the policy version currently being trained and that recorded version, i.e., the number of policy updates applied since the token was generated.
A token is \emph{fresh} if its staleness is at most $s_{\max}$. 
A trajectory is admitted for training only if its fresh-token ratio reaches a threshold $\rho$, and within admitted trajectories, the stale prefix tokens are masked out of the loss (Figure~\ref{fig:fresh_gateway}).
No over-stale token contributes gradient, while long, expensive trajectories are salvaged rather than discarded.

\begin{figure}[htbp]
\centering
\includegraphics[width=0.8\linewidth]{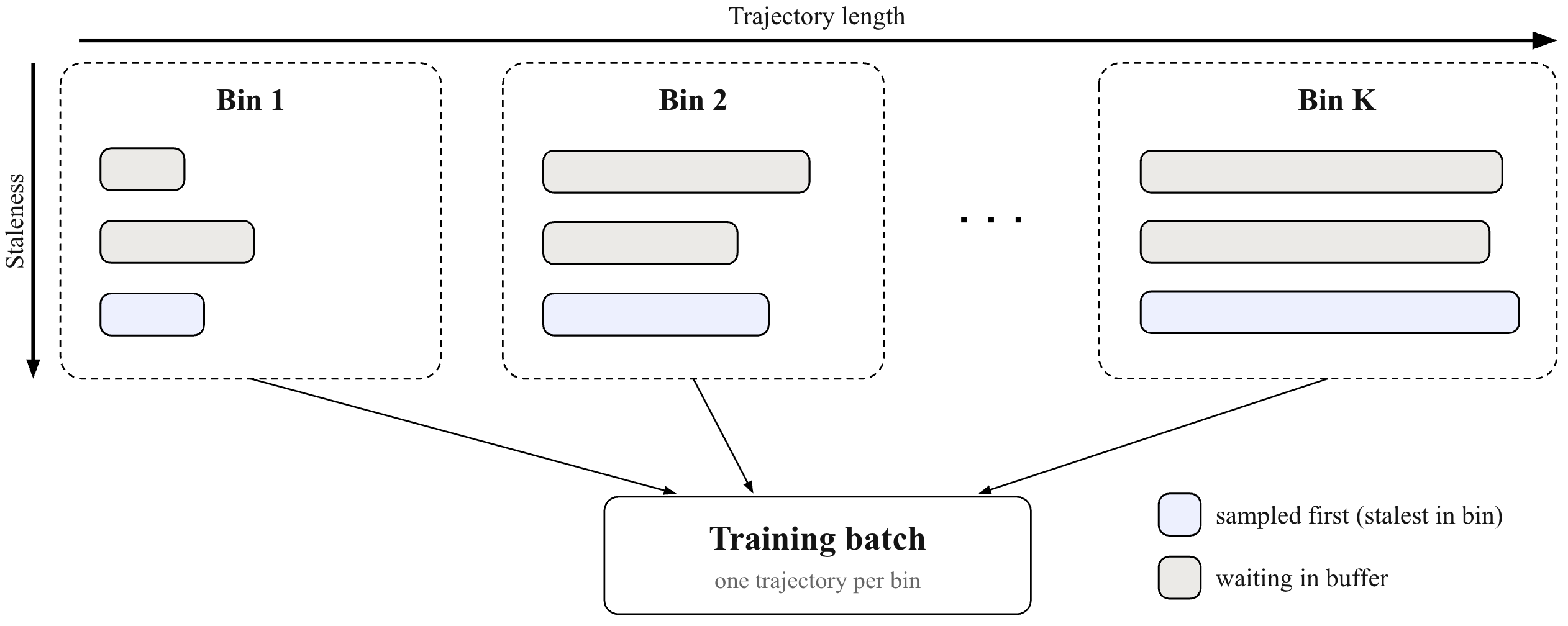}
\caption{Length- and staleness-aware batch sampling. Buffered trajectories are partitioned by length into bins $1$ through $K$; each training batch draws one trajectory per bin, spanning the full length spectrum, and within each bin the stalest trajectory (blue) is sampled first, ahead of those waiting in the buffer (gray).}
\label{fig:binned_sampling}
\end{figure}

\paragraph{Length- and staleness-aware batch sampling.} A subtle failure mode of fully asynchronous training is length bias at startup: short trajectories finish first, so FIFO sampling would train on a length-skewed distribution and risk spurious correlations between response length and reward. 
Before the first optimizer step, we pre-generate $N$ times the training batch size in trajectories, so that long rollouts have the wall-clock time to populate the buffer.
At each step, the buffered trajectories are sorted by length and partitioned into $\lfloor |\mathrm{buffer}| / |\mathrm{batch}| \rfloor$ bins, and one trajectory is drawn per bin, so every batch spans the full length spectrum. Within each bin, the stalest trajectories are drawn first, so samples closest to eviction are consumed before they expire, maximizing data utilization (Figure~\ref{fig:binned_sampling}).

\paragraph{Direct double-sided importance sampling.} The rollout policy may be updated multiple times during the generation of a single trajectory, so the drift between rollout and training policies can be far larger than in the standard on-policy setting.
We adopt the direct double-sided importance sampling of \citet{zeng2026glm5}: the token-level importance ratio is defined directly against the rollout policy from the log-probabilities recorded at generation time, and tokens whose ratio falls outside a double-sided trust region are excluded from the update entirely rather than clipped. 
Retaining token-level ratios as in GRPO~\citep{shao2024deepseekmath}, rather than the sequence-level ratios of GSPO~\citep{zheng2025group}, follows from the design itself: different tokens within one trajectory may originate from different policy versions, so off-policyness must be controlled at the granularity at which it arises.
This trust-region mask is applied orthogonally to the staleness mask above: a token contributes gradient only if it is both fresh and within the trust region.

\paragraph{Environment-failure filtering and group-size repair.} In sandboxed agentic environments, a trajectory can receive a poor reward through no fault of the policy (file-system errors, container crashes, and other infrastructure failures), and training on such spurious negative rewards injects noise into the advantage estimates.
We log a failure reason for every unsuccessful trajectory and exclude those attributable to environment collapse. 
Since dropping samples perturbs the GRPO group size, groups are repaired: if more than half of a group's rollouts remain valid, the valid samples are duplicated to restore the original group size; otherwise the entire group is discarded.

\subsubsection{Multi-Teacher On-Policy Distillation}\label{sec:mopd}

Post-training separates \emph{capability cultivation} from \emph{capability integration}: the Specialist stage grows a pool of domain experts independently, and MOPD folds them into one deployable model by letting the student learn from dense, full-vocabulary teacher signal on its own rollouts, all running on the fully asynchronous infrastructure of Section~\ref{sec:async-rl}.

\paragraph{From twelve specialists to one model.} 
Each target capability is cultivated independently --- domain SFT followed by large-scale RL with verifiable rewards --- yielding a specialist that peaks on its own domain but is neither balanced nor deployable alone. 
The twelve specialists fall into three capability families: \emph{reasoning} (math, STEM, code), \emph{agents \& tools} (coding, general tool use, single-workspace, multi-workspace, search), and \emph{preference \& alignment} (instruction following, human preference, safety, abstention). 
Consolidation must merge them without trading one domain's peak for another's. 
The naive route --- supervised fine-tuning the student on teacher-generated traces --- is off-policy: the student receives no supervision on the error-compounded states its own sampling drifts into (exposure bias), and a consolidated model asked to operate across every domain at once is exactly what exposes this mismatch most. 
MOPD closes the gap in the spirit of interactive imitation learning~\citep{ross2011dagger}: the student generates its own trajectories, the routed teacher scores those tokens against its domain distribution, and the student is pulled toward the teacher along the states it actually visits --- dense, token-level supervision on the student's own distribution, far more sample-efficient than a sparse outcome reward.

\paragraph{The MOPD setting.} Each prompt is routed to exactly one of the twelve specialist teachers, and the student minimizes the per-position reverse KL to the routed teacher along its own trajectories. Every per-position term is computed in closed form over the full vocabulary rather than estimated from the sampled token. 
This distillation KL is the entire training objective, and its mode-seeking character concentrates the student on the teacher's high-confidence continuations --- exactly what capability transfer wants. 
The setting departs from prior on-policy-distillation consolidation recipes~\citep{deepseekai2026deepseekv4, nvidia2026nemotron, xiaomi2026mimo, ma2026mopd} in two ways. 
First, the exact full-vocabulary computation removes the single-draw estimator variance and stays well-defined on student-drifted prefixes the teacher rarely visits --- precisely where consolidation is hardest.
Second, no verifiable outcome reward is added on top of the KL: the $G$ rollouts per prompt that a group-relative advantage would require disappear, so consolidation runs at one rollout per prompt, with no $\lambda$ to balance and no reward-hacking surface. The rollout pump, the throughput bottleneck, shrinks by a factor of $G$, and at the same generation budget every optimizer step spans $G$ times more distinct prompts across all twelve domains.
Each specialist was itself cultivated with verifiable-reward RL, so its reward is already implicit in its distribution and re-adding it is largely redundant. The headroom to surpass the teacher that a reward term would buy is deliberately traded away: the goal of consolidation is to match twelve already-strong specialists, not to out-train them.
In practice, the KL-only objective descends stably: entropy contracts without collapsing, and training stays effectively on-policy.

\paragraph{Scaling full-vocabulary MOPD to 250B.} A full-vocabulary, multi-teacher signal is trivial to write down and expensive to serve; four systems problems stand between the objective and a run at this scale. 
\begin{itemize}
    \item \emph{Teacher placement cost.} A colocated placement hot-swaps the CPU-pinned teacher pool onto the actor GPUs each step (offload the student, forward the teacher, onload, train), so the teacher forward sits serially on the training critical path. We instead give the pool dedicated nodes, collapsing the per-step teacher tax toward zero at the price of the extra nodes.
    \item \emph{Dense signal without the $[L \times V]$ blow-up.} The exact objective wants the teacher's distribution over all 196{,}608 ($V$) tokens at every response position ($L$), on the order of 26\,GB per microbatch in fp32. The teacher therefore ships only its pre-\texttt{lm\_head} hidden states $[L \times H]$ ($H = 4{,}096$; $48\times$ smaller), and the student rebuilds and reduces the logits one 1{,}024-token tile at a time with activation checkpointing, keeping the GPU peak to a single $[\mathrm{tile} \times V]$ buffer ($\approx$1.1\,GB) without ever materializing the full matrix or approximating the objective.
    \item \emph{Transport cost.} Because each sample routes to exactly one teacher, total teacher compute is $O(\mathrm{batch})$, independent of the number of teachers. Each teacher caches results keyed by global sample index, so the student data-parallel width can grow with no teacher recompute. Under context parallelism, the teacher's CP rank $r$ already holds exactly the shard student CP rank $r$ needs, so the transport skips the CP all-gather and ships only the matching shard.
    \item \emph{Teacher pool management.} Twelve 250B teachers cannot co-reside on the accelerators, so the pool keeps one GPU-resident model plus twelve CPU parameter snapshots and swaps the routed teacher into the single GPU slot on demand. Micro-batches are packed by routed teacher, so consecutive micro-batches usually hit the resident model and the swap cost is amortized across the packed run.
\end{itemize}

\section{Evaluation}\label{sec:evaluation}

We evaluate \modelname in three parts. Section~\ref{sec:eval-setup} describes the benchmark suites and the measurement protocol. Sections~\ref{sec:eval-english} and~\ref{sec:eval-korean} present the headline comparison: on English benchmarks \modelname is competitive with the strongest open-weight models, and on Korean benchmarks it records the highest average among all models compared, including fast-tier closed APIs. Section~\ref{sec:kogdpval} then examines the sovereign target scenario in depth through Ko-GDPval, an agentic benchmark for Korean officework.

\subsection{Setup}\label{sec:eval-setup}

\paragraph{English suite.} The English benchmarks are organized into the three groups shown in Table~\ref{tab:eng_benchmarks}. \emph{Knowledge \& reasoning} covers MMLU-Pro~\citep{wang2024mmlu}, GPQA-Diamond~\citep{rein2024gpqa}, HLE without tools~\citep{phan2025humanity}, LiveCodeBench v6~\citep{jain2024livecodebench}, ArtifactsBench~\citep{zhang2025artifactsbenchbridgingvisualinteractivegap}, and the competition-math sets HMMT2602 and AIME2026 from MathArena~\citep{balunovic_srimatharena_2025}. \emph{Instruction following \& long context} covers Multi-Challenge~\citep{sirdeshmukh2025multichallengerealisticmultiturnconversation}, IFBench~\citep{pyatkin2025generalizing}, and AA-LCR~\citep{artificialanalysis2025lcr}. \emph{Agents} covers SWE-Bench Verified~\citep{jimenez2023swebench}, Terminal Bench Hard~\citep{merrill2026terminalbench}, APEX-Agents~\citep{vidgen2026apexagents}, MCP-Atlas~\citep{bandi2026mcpatlaslargescalebenchmarktooluse}, the $\tau^3$ banking domain of the $\tau$-bench suite~\citep{yao2024tau, barres2025tau}, and GDPval-AA v2~\citep{patwardhan2025gdpval}, reported as an ELO rating measured on 30 June 2026.

\paragraph{Korean suite.} The Korean benchmarks (Table~\ref{tab:korean_benchmarks}) span five groups: \emph{general knowledge}, measured by KMMLU-Pro~\citep{hong2025kmmlupro}; \emph{Korean language \& culture}, by CLIcK~\citep{kim2024click} and HaeRae~\citep{son2024hae}; \emph{mathematical reasoning}, by Ko-AIME 2025 and HRM8K~\citep{ko2025understand}; \emph{professional domains}, by KBank-MMLU (finance), KBL~\citep{kim2024kbl} (law), and KorMedMCQA~\citep{kweon2024kormedmcqa} (medicine); and \emph{officework agents}, by Ko-GDPval (Section~\ref{sec:kogdpval}). Ko-AIME, KBank-MMLU, and Ko-GDPval are in-house benchmarks; the rest are public.

\paragraph{Baselines and protocol.} We compare against strong open-weight models under 320B total parameters, the scale band \modelname is designed for: DeepSeek-V4-Flash (284B-A13B), MiMo-V2.5 (310B-A15B), Command A+ (218B-A25B), and Mistral Medium 3.5 (128B dense), together with the previous generation Solar Open 1 (102B-A12B). On the Korean suite we additionally compare two fast-tier closed APIs: Claude Haiku 4.5 and GPT-5.4 mini. All numbers are measured with our internal evaluation harness under identical per-benchmark settings; reasoning-effort settings are noted in the table headers where applicable.

\begin{table}[t]
\centering
\caption{English benchmark performance. All numbers are measured with our internal evaluation harness. Model sizes are shown as total parameters--activated parameters (A) for MoE models.}
\label{tab:eng_benchmarks}
\tabcolsep=4pt
\resizebox{\textwidth}{!}{%
\begin{tabular}{@{}c l >{\columncolor{SolarViolet!10}}c c c c c c@{}}
\toprule
 & \textbf{Benchmark} & \textbf{\modelname} & \textbf{Solar Open 1} & \textbf{Command A+} & \textbf{Mistral Medium 3.5} & \textbf{MiMo-V2.5} & \textbf{DeepSeek-V4-Flash} \\
 & & \textbf{{\footnotesize 250B-A15B}} & \textbf{{\footnotesize 102B-A12B}} & \textbf{{\footnotesize 218B-A25B}} & \textbf{{\footnotesize 128B dense, high}} & \textbf{{\footnotesize 310B-A15B}} & \textbf{{\footnotesize 284B-A13B, max}} \\
\midrule
\multirow{7}{*}{\rotatebox[origin=c]{90}{\footnotesize\sffamily Know. \& Reasoning}}
 & MMLU-Pro & 86.2 & 80.4 & 79.0 & 81.2 & 84.6 & 85.9 \\
 & GPQA-Diamond & 86.3 & 66.2 & 75.6 & 77.5 & 83.0 & 88.9 \\
 & HLE (w/o tools) & 28.8 & 11.5 & 11.4 & 12.8 & 24.3 & 32.3 \\
 & LiveCodeBench (v6) & 92.4 & 56.5 & 86.1 & 84.9 & 89.1 & 92.3 \\
 & ArtifactsBench & 55.9 & 43.4 & 42.8 & 49.8 & 59.3 & 61.0 \\
 & HMMT2602 & 93.9 & 68.9 & 73.5 & 62.9 & 61.4 & 94.7 \\
 & AIME2026 & 95.7 & 87.7 & 96.0 & 89.0 & 92.3 & 97.0 \\
\midrule
\multirow{3}{*}{\rotatebox[origin=c]{90}{\footnotesize\sffamily IF / Long}}
 & Multi-Challenge & 61.0 & 40.5 & 45.8 & 49.8 & 39.0 & 62.0 \\
 & IFBench & 80.0 & 57.7 & 73.9 & 69.0 & 67.1 & 80.3 \\
 & AA-LCR & 62.3 & 36.0 & 46.0 & 61.0 & 62.7 & 63.7 \\
\midrule
\multirow{6}{*}{\rotatebox[origin=c]{90}{\footnotesize\sffamily Agent}}
 & SWE-Bench Verified & 70.4 & 15.4 & 14.4 & 69.6 & 73.0 & 73.8 \\
 & Terminal Bench Hard & 28.3 & 2.3 & 25.0 & 33.3 & 41.7 & 34.1 \\
 & APEX-Agents & 16.6 & 2.4 & 1.6 & 6.1 & 13.4 & 13.2 \\
 & MCP-Atlas & 58.2 & 34.4 & 27.2 & 30.7 & 63.9 & 58.2 \\
 & $\tau^3$ (banking) & 19.6 & 7.4 & 5.8 & 5.8 & 8.7 & 22.3 \\
 & GDPval-AA v2 (ELO) & 1128 & -- & 712 & 929 & 1145 & 1187 \\
\bottomrule
\end{tabular}%
}
\end{table}

\subsection{English Results}\label{sec:eval-english}

Table~\ref{tab:eng_benchmarks} shows that \modelname is competitive with the strongest open-weight baselines, DeepSeek-V4-Flash and MiMo-V2.5, both larger in total parameters. On knowledge and reasoning it posts the best scores in the comparison on MMLU-Pro (86.2) and LiveCodeBench v6 (92.4), and on GPQA-Diamond (86.3), HLE (28.8), HMMT2602 (93.9), and Multi-Challenge (61.0) it is second only to DeepSeek-V4-Flash while clearly ahead of MiMo-V2.5. Instruction following is effectively tied at the top (IFBench 80.0 vs.\ 80.3; AA-LCR 62.3 vs.\ 63.7), and ArtifactsBench (55.9) is the one knowledge-group benchmark where both frontier baselines stay ahead.

On the agent group the picture is competitive with a localized gap. \modelname records the best APEX-Agents score in the comparison (16.6), matches DeepSeek-V4-Flash on MCP-Atlas (58.2 for both; MiMo-V2.5 leads at 63.9), and is second on the $\tau^3$ banking domain (19.6 vs.\ 22.3). On the GDPval-AA v2 ELO it places third at 1128, behind DeepSeek-V4-Flash (1187) and within 17 points of MiMo-V2.5 (1145) --- a margin that corresponds to a nearly even expected head-to-headd win rate against a model 60B larger in total parameters. The remaining gap concentrates in repository- and terminal-level software engineering: SWE-Bench Verified 70.4 versus 73.0--73.8 for the two frontier baselines, and Terminal Bench Hard 28.3 versus MiMo-V2.5's 41.7.

The generational jump is largest exactly where Solar Open 1 fell short (Section~\ref{sec:introduction}): agentic execution (SWE-Bench Verified 15.4$\rightarrow$70.4, MCP-Atlas 34.4$\rightarrow$58.2, APEX-Agents 2.4$\rightarrow$16.6) and frontier-level reasoning (GPQA-Diamond 66.2$\rightarrow$86.3, HLE 11.5$\rightarrow$28.8, HMMT2602 68.9$\rightarrow$93.9). 

\begin{table}[t]
\centering
\caption{Korean benchmark performance. All numbers are measured with our internal evaluation harness; Avg.\ is the mean over the listed benchmarks. }
\label{tab:korean_benchmarks}
\tabcolsep=4pt
\resizebox{\textwidth}{!}{%
\begin{tabular}{@{}l >{\columncolor{SolarViolet!10}}c c c c c c@{}}
\toprule
\textbf{Benchmark} & \textbf{\modelname} & \textbf{Solar Open 1} & \textbf{MiMo-V2.5} & \textbf{DeepSeek-V4-Flash} & \textbf{Claude Haiku 4.5} & \textbf{GPT-5.4 mini} \\
 & \textbf{{\footnotesize 250B-A15B}} & \textbf{{\footnotesize 102B-A12B}} & \textbf{{\footnotesize 310B-A15B}} & \textbf{{\footnotesize 284B-A13B, max}} & \textbf{{\footnotesize closed}} & \textbf{{\footnotesize closed}} \\
\midrule
KMMLU-Pro & 78.4 & 64.0 & 69.1 & 78.9 & 67.9 & 78.1 \\
CLIcK & 90.7 & 78.9 & 78.4 & 89.2 & 53.5 & 89.6 \\
HAE-RAE v1.1 & 73.8 & 73.3 & 61.7 & 73.1 & 38.5 & 69.4 \\
Ko-AIME'25$\dagger$ & 97.7 & 80.0 & 88.0 & 98.0 & 81.7 & 90.7 \\
HRM8K & 92.2 & 87.6 & 90.7 & 93.4 & 90.6 & 91.3 \\
KBank-MMLU$\dagger$ & 80.8 & 65.5 & 71.0 & 79.5 & 68.9 & 79.0 \\
KBL & 75.5 & 65.5 & 69.8 & 72.8 & 69.9 & 75.3 \\
KorMedMCQA & 93.0 & 84.4 & 87.7 & 94.1 & 87.0 & 94.2 \\
Ko-GDPval$\dagger$ & 86.8 & 3.4 & 81.0 & 85.0 & 68.3 & 59.4 \\
\midrule
Avg. & 85.4 & 67.0 & 77.5 & 84.9 & 69.6 & 80.8 \\
\bottomrule
\end{tabular}%
}
\vspace{5pt}
\raggedright
{\footnotesize $\dagger$ in-house benchmarks.}
\end{table}

\subsection{Korean Results}\label{sec:eval-korean}

On the Korean suite (Table~\ref{tab:korean_benchmarks}), \modelname records the highest average of all six models compared (85.4), ahead of DeepSeek-V4-Flash (84.9) and of both fast-tier closed APIs --- GPT-5.4 mini (80.8) and Claude Haiku 4.5 (69.6). The lead is broad-based across the five groups: it is first on CLIcK (90.7) in language \& culture, first on two of the three professional domains --- KBank-MMLU (80.8) and KBL (75.5) --- and within a point of the lead on KMMLU-Pro (78.4) and Ko-AIME (97.7). Its decisive edge is Ko-GDPval (86.8), the deliverable-producing officework task, where it leads DeepSeek-V4-Flash by 1.8 points and every other model by more than 5.

The generational gain over Solar Open 1 is +18.4 points on average (67.0 $\rightarrow$ 85.4) and on the agentic axis it is qualitative: the previous generation effectively could not perform deliverable-producing officework (Ko-GDPval 3.4), whereas \modelname is the strongest model measured on that task. Together with the English results, this delivers the design goal set out in Section~\ref{sec:introduction} --- to be competitive with the global open frontier in English, and leadership in the sovereign scenario of Korean knowledge work --- which we attribute to the Korean-efficient tokenizer (Section~\ref{sec:tokenizer}), the Korean data curation of pre-training, and the officework agent scenarios of Section~\ref{sec:officework}. The next section examines that scenario in depth. 

\begin{figure}[htbp]
\centering
\includegraphics[width=0.9\linewidth]{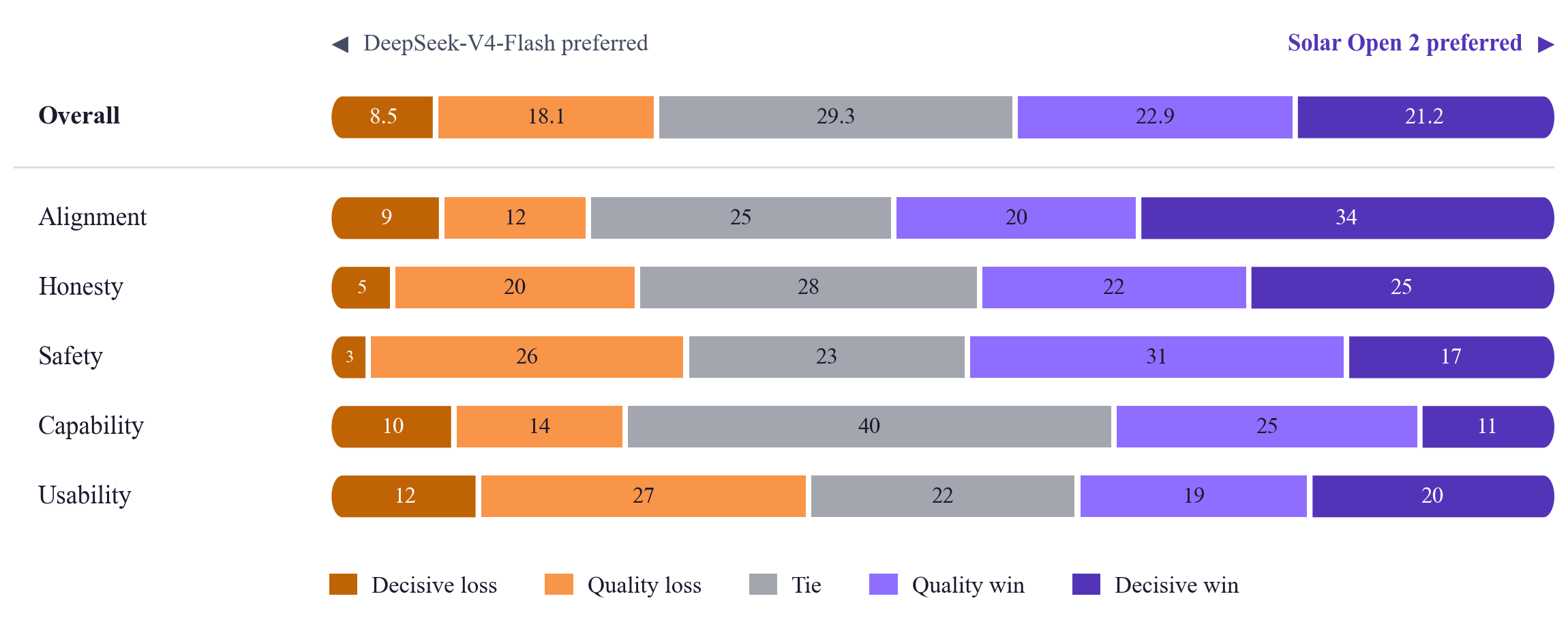}
\caption{Blind pairwise qualitative evaluation on Korean conversations (\modelname vs.\ DeepSeek-V4-Flash). Each bar is that aspect's outcome distribution in percent, centred on the tie segment, with rows ordered by net margin.}
\label{fig:qualitative_pairwise}
\end{figure}

We also ran a blind pairwise qualitative evaluation against DeepSeek-V4-Flash on 835 Korean conversation turns. \modelname is preferred on 44.1\% of turns versus 26.6\%, and wins decisively --- the other answer got a fact wrong or came out garbled, which no other strength can make up for --- on 21.2\% versus 8.5\% (Figure~\ref{fig:qualitative_pairwise}). Alignment, honesty, and safety lead by the widest margins because on these axes the correct answer is culturally located --- which perspective a contested Korean issue deserves, which Korean claim may be asserted rather than hedged, which request counts as harmful under Korean norms --- and a model can be fluent in Korean while still defaulting to a foreign frame on all three. Usability comes out even, and its losses are almost entirely quality rather than decisive ones (27\% vs.\ 12\%) --- a matter of phrasing preference rather than of anything a Korean-specialized model encodes differently.

\subsection{Korean Officework Agent Performance: Ko-GDPval}\label{sec:kogdpval}

Ko-GDPval is an in-house Korean officework agentic benchmark that inherits the paradigm of GDPval~\citep{patwardhan2025gdpval} --- evaluating ``economically valuable real-world work'' --- and roots it in Korean industry, law, and business practice. It comprises 170 tasks, all in Korean, spanning 11 industry domains in which economically valuable, deliverable-centric knowledge work is concentrated. Each task is a workplace scenario from one of 58 occupations --- lawyers, CPAs, infection-control specialists, project-finance underwriters, and more --- with the field's actual statutes, formulas, and procedures reflected in the instruction and in the grading rubric.

\begin{table}[t!]
\small
\centering
\caption{Ko-GDPval overall results. Score is the overall benchmark score; the remaining columns break performance down by grading dimension. }
\label{tab:kogdpval_results}
\tabcolsep=4pt
\resizebox{\textwidth}{!}{%
\begin{tabular}{@{}l l | c |  c c c c c c c@{}}
\toprule
\textbf{Model} & \textbf{Size} & \textbf{Score} & \textbf{Data} & \textbf{Content} & \textbf{Structure} & \textbf{Quality} & \textbf{Value Format} & \textbf{Constraint} & \textbf{File Format} \\
\midrule
DeepSeek-V4-Pro & 1.6T & 86.9 & 82.1 & 90.5 & 96.6 & 84.2 & 81.0 & 88.2 & 99.4 \\
\textbf{\modelname} & 250B & \textbf{86.8} & 83.5 & 89.1 & 96.5 & 79.2 & 82.0 & 86.7 & 99.8 \\
DeepSeek-V4-Flash & 284B & 85.0 & 81.0 & 87.6 & 95.9 & 78.2 & 78.3 & 86.9 & 100.0 \\
MiMo-V2.5-Pro & 1T & 84.6 & 80.6 & 87.7 & 95.2 & 75.8 & 78.5 & 86.4 & 99.2 \\
MiMo-V2.5 & 310B & 81.0 & 76.2 & 84.0 & 93.4 & 72.6 & 73.1 & 84.8 & 97.9 \\
Claude Haiku 4.5 & closed & 68.3 & 60.6 & 71.3 & 88.6 & 56.2 & 59.7 & 78.2 & 99.8 \\
GPT-5.4 mini & closed & 59.4 & 54.4 & 54.6 & 85.7 & 36.8 & 43.7 & 83.1 & 97.5 \\
\bottomrule
\end{tabular}%
}
\end{table}

\modelname stands just behind the strongest model compared, DeepSeek-V4-Pro (86.9) --- a 1.6T-parameter model more than six times its size --- and ahead of every other model measured, including the 1T-parameter MiMo-V2.5-Pro (84.6) and DeepSeek-V4-Flash (85.0; Table~\ref{tab:kogdpval_results}). 
A more fine-grained analysis by rubric type shows that \modelname's largest gap relative to DeepSeek-V4-Pro lies in Quality dimension (79.2 vs.\ 84.2), while it leads on Data (83.5 vs.\ 82.1) and Value Format (82.0 vs.\ 81.0). In the sovereign scenario of Korean officework, the native-language-specialized 250B model delivers the officework capability of frontier-scale models at a fraction of their size and leads the fast-tier frontier APIs it is built to compete with.

\begin{figure}[t!]
\centering
\includegraphics[width=.9\linewidth]{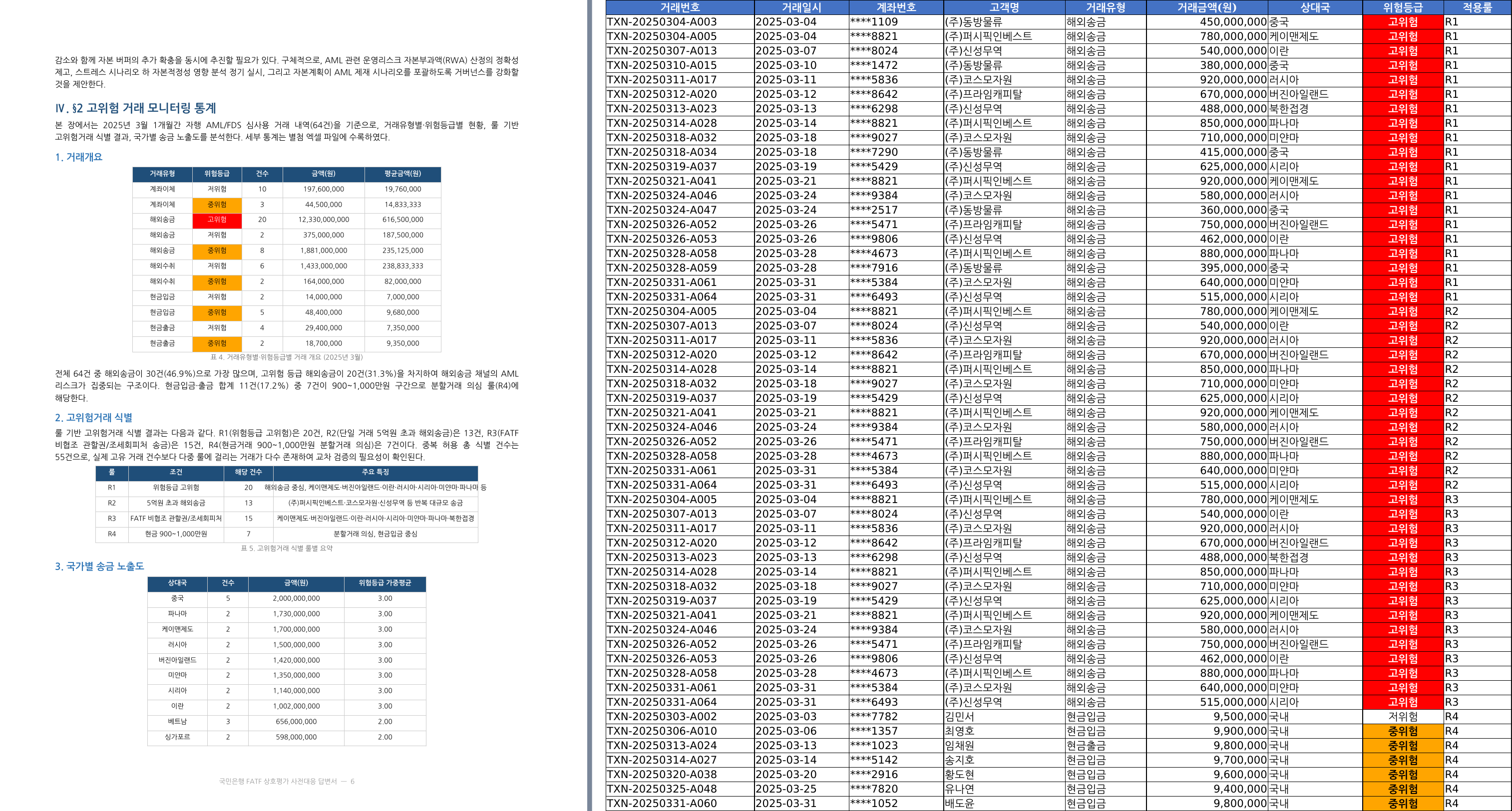}
\caption{FATF mutual-evaluation pre-response, delivered in two mutually consistent formats. Left: the regulator-facing response report (PDF), summarizing transactions by risk grade and country exposure. Right: the monitoring workbook (xlsx) listing all 55 rule-flagged (R1–R4) transactions.}
\label{fig:kogdpval_qual1}
\end{figure}

\begin{figure}[t!]
\centering
\includegraphics[width=.9\linewidth]{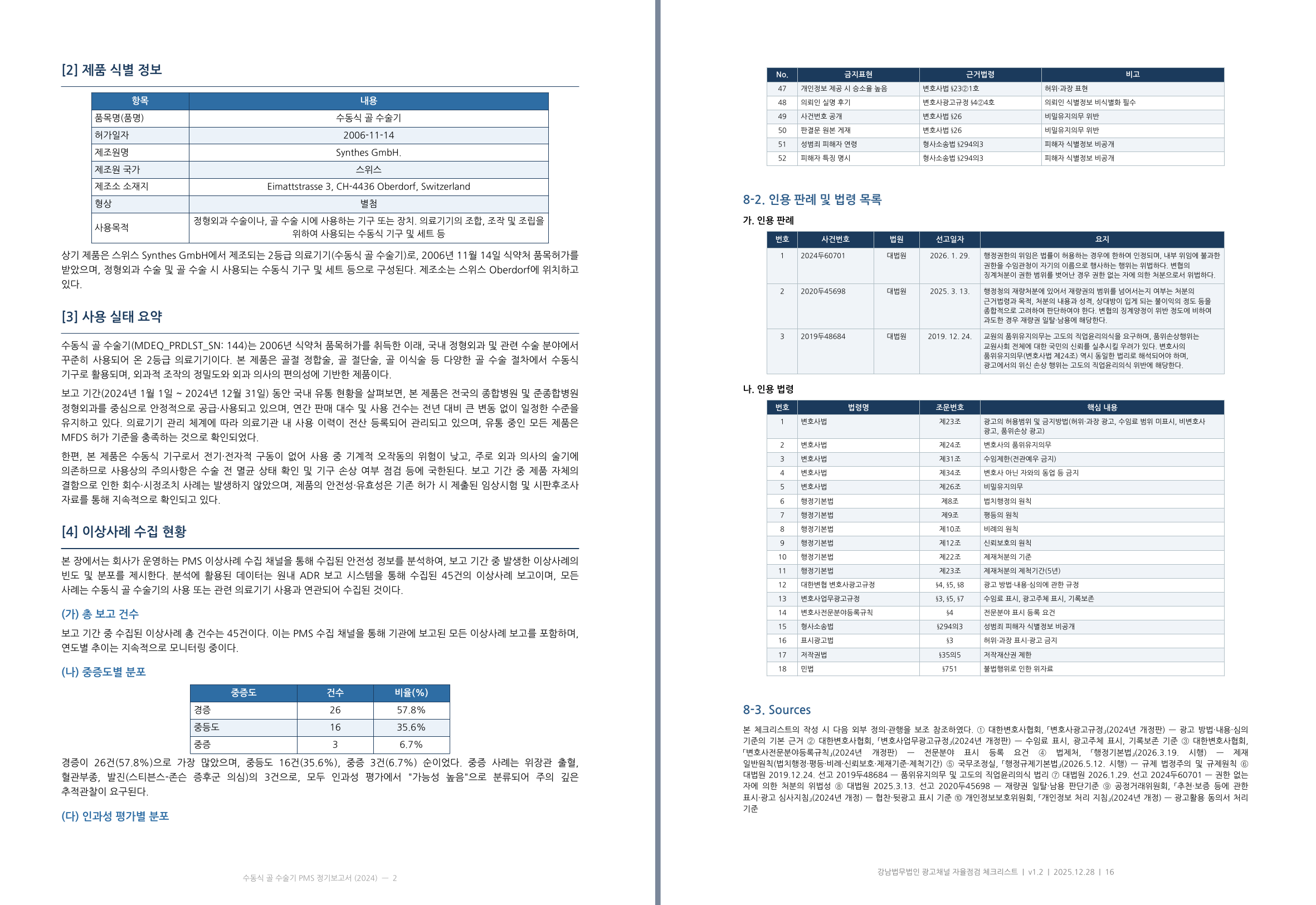}
\caption{Source fidelity and cited grounds across two regulatory-document tasks. Left: a medical-device PMS periodic report, with product identification and adverse-event severity tallies. Right: a law firm's advertising self-inspection, with its cited-precedent table and statute index.}
\label{fig:kogdpval_qual2}
\end{figure}

Quantitative scores alone do not reveal what these tasks demand or what the model actually produces, so we inspected the submitted deliverables directly. 
Figures~\ref{fig:kogdpval_qual1} and~\ref{fig:kogdpval_qual2} show two representative examples, each completed autonomously by \modelname in the sandbox from the reference files alone. The first is a FATF mutual-evaluation pre-response task (compliance officer, finance \& insurance), hard for two reasons: dozens of ledger transactions must be screened precisely against four AML rules (R1--R4) and aggregated into country-level risk exposure, and the result must be delivered twice --- as a regulator-facing pdf report and as an xlsx monitoring workbook --- with the two kept consistent. \modelname completes both in one pass, the report restating the workbook's analysis in document form (Figure~\ref{fig:kogdpval_qual1}). The second pairs two regulatory-document tasks --- a medical-device PMS periodic report, and a law firm's advertising self-inspection --- whose difficulty is source fidelity and traceable grounds: identifications and counts must be transcribed exactly from the references, and every judgment must cite the statute and article it rests on. \modelname's deliverables hold to this standard --- verbatim product identification, severity tallies that sum exactly to the source counts, and precedent tables and statute indexes that make each checklist judgment traceable (Figure~\ref{fig:kogdpval_qual2}). Additional deliverable examples are provided in Section~\ref{sec:appendix:officework}.

\section{Conclusion}\label{sec:conclusion}

We presented \modelname, a 250B-A15B Mixture-of-Experts model built as a Korean sovereign AI for long-horizon agents.
Three threads run through this report. 
Architecturally, a hybrid attention stack --- three linear-attention layers per softmax layer, with NoPE and the negative-eigenvalue extension --- extends the context window to 1M tokens at about a quarter of the memory and computation of an all-softmax design.
In pre-training, a selective weight transfer initializes about 2.3\% of the model's parameters from Solar Open 1, the portion that survives the architectural change, and a quality- and rarity-aware curation pipeline refines a 20T-token pool into the 10T mixture consumed by a staged curriculum with checkpoint merging.
In post-training, purpose-built agent scenarios spanning conversational tool use, coding, and officework feed a fully asynchronous RL system. Multi-teacher on-policy distillation (MOPD) then folds twelve domain specialists into one model.
The result is a model competitive with the strongest comparably sized open-weight models on English benchmarks and first on average across the Korean suite, ahead of even fast-tier closed APIs. On Ko-GDPval --- the sovereign target scenario of deliverable-producing Korean officework --- \modelname essentially matches DeepSeek-V4-Pro (1.6T) at less than a sixth of its size, a capability that effectively emerges in this generation, rising from Solar Open 1's 3.4 to 86.8.

The remaining gaps are localized and point to the next steps: repository- and terminal-level software engineering on the English agent suite, and fine-grained numerical precision in officework deliverables. Both call for stronger verification and self-checking behaviors in agent training.
Beyond these gaps, we view the Solar Open recipe --- tokenizer efficiency, cross-generation selective weight transfer, data curation optimizing value per token, and scenario-driven agent post-training with specialist consolidation --- as a reusable path toward sovereign models for other underserved languages.

\section*{Acknowledgments}
Special thanks to our colleagues at Upstage — Eunbi Cho, Eunhae Choo, Jisun Kim, Sangwon Joo, and Seungwon Cheon — whose tireless support and expertise were instrumental in the development and release of \modelname.

We are deeply grateful to the members of our consortium for their
contributions throughout the project: Lablup Inc. for building and operating
the GPU clusters and large-scale training infrastructure, Nota Inc. for
model quantization enabling efficient inference and serving, and Flitto for
constructing large-scale training datasets. We also thank our domain and ecosystem partners — Alfred Co., Ltd., Allganize Korea Co., Ltd., Asteromorph Inc., AXZ Corp.,
Channel Corp., clush Inc., Dalpha Inc., DAY 1 COMPANY,
Finda, Inc., HyperAccel Co., Ltd., Korea Electronics Technology Institute, Korea Financial Telecommunications and Clearings Institute (KFTC),
Law\&Company Co., Ltd., MakinaRocks Co., Ltd., RLWRLD Inc. and VUNO Inc. — for contributing domain expertise and evaluation,
and for supporting the adoption and dissemination of the model across
industries. Finally, we thank our academic partners, New York University, Stanford University, KAIST and Sogang University for their
research collaboration throughout the project.

This research was conducted as part of the Sovereign AI Foundation Model Project (GPU Track), organized by the Ministry of Science and ICT (MSIT) and supported by the National IT Industry Promotion Agency (NIPA), South Korea (PJT-26-010017).
This work was also conducted as part of the Sovereign AI Foundation Model Project (Data Track), organized by MSIT and supported by the National Information Society Agency (NIA), South Korea (Grant No. 2026-AIData-WII03).
In addition, this research was supported by the MSIT, South Korea, under the Top-Tier AI Global HRD invitation program (RS-2025-25461932) supervised by the IITP (Institute for Information \& Communications Technology Planning \& Evaluation).

\bibliographystyle{plainnat}
\bibliography{references}

\appendix
\newpage
\section{Appendix}\label{sec:appendix}

\subsection{Author List}
Every author is affiliated with Upstage, South Korea, unless specified otherwise. Authors in each group made equal contributions. 
\paragraph{Core Contributors}
Sungrae Park,  
Sanghoon Kim,  
Gyoungjin Gim, 
Jungho Cho,    
Hyunwoong Ko,  
Minbyul Jeong, 
Minjeong Kim,  
Keunwoo Choi,  
Chaehun Shin, 
Chanwoong Yoon, 
Dongjun Kim, 
Eunwon Kim, 
Gyungin Shin, 
Hyeonju Lee, 
Hyungkyu Kang, 
Inseo Song, 
Jisu Bae, 
Jiyoon Han, 
Jiyun Lee, 
Joonkee Kim, 
Junyeop Lee, 
Mikyoung Cha, 
Sangwon Yu, 
Sehwan Joo, 
Seokyoon Kang, 
Seonghoon Yang, 
Seung Shin, 
Seunghyun Lee, 
Seungseop Lim, 
Seungyoun Shin, 
Sukyung Lee, 
Taegyeong Eo, 
Taehwan Oh, 
Taewhoo Lee, 
Wonho Song, 
Wonjun Oh, 
Wonseok Hwang (University of Seoul), 
Yunsu Kim, 
Yura Shim

\paragraph{Contributors}
Hwalsuk Lee,  
Sunghun Kim,  
Du-Seong Chang (Department of Artificial Intelligence, Sogang University),
Kyunghyun Cho (New York University),
Seungju Han (Stanford),
Yejin Choi (Stanford),
Junsuk Choe (Sogang University),
Hwaran Lee (Sogang University),
Minjeong Ban (KAIST),
Hwanjun Song (KAIST),
Jae-Gil Lee (KAIST),
KyungTae Lim (KAIST),
Alice Oh (KAIST)

\subsection{Korean Officework Agent Task Example}\label{sec:appendix:officework}

\newtcolorbox{taskprompt}{%
  enhanced jigsaw, breakable,
  colback=black!3, colframe=black!60,
  boxrule=0.5pt, arc=1.2pt,
  left=8pt, right=8pt, top=6pt, bottom=7pt,
  title={Task prompt},
  fonttitle=\small\bfseries, coltitle=black, colbacktitle=black!10,
  fontupper=\small,
  before skip=8pt, after skip=10pt}
\newcommand{\promptsec}[1]{\par\medskip\noindent\textbf{#1}\par\smallskip}
\newlist{pitemize}{itemize}{2}
\setlist[pitemize]{label=\textbullet, nosep, leftmargin=1.35em, topsep=2pt}
\newlist{penum}{enumerate}{1}
\setlist[penum]{label=\arabic*., nosep, leftmargin=1.6em, topsep=2pt}
\newcommand{\showcasemeta}[3]{\noindent{\small\itshape Task \texttt{#1}~\textperiodcentered~ #2 rubric items ~\textperiodcentered~ rubric score #3}\par}

\subsubsection{New-drug reimbursement listing application (PDF)}
\label{app:showcase-s2}

\begin{taskprompt}
당신은 국내 중견 제약사 시장접근(Market Access)팀의 약가·급여 등재 담당
선임연구원입니다. 자체 개발한 제2형 당뇨병 치료 신약(코드명: KMD-217, 일반명:
가칭 ‘Kameglitin’)의 식약처 품목허가를 앞두고 건강보험심사평가원(HIRA)에
요양급여 등재를 신청해야 하며, 등재 신청 근거 자료를 통합한 제출용 신청서를
작성하는 것이 이번 업무입니다.

\promptsec{배경}
\begin{pitemize}
  \item 신약명(가칭): Kameglitin (KMD-217), 제2형 당뇨병 단독요법
  \item 활성 대조약제: Sitagliptin 100mg (DPP-4 억제제)
  \item 임상 2상 결과 기반
  \item 신청 기준일: 2025-12-01 / 제출 마감일: 2026-01-30
\end{pitemize}

\promptsec{첨부 자료}
\begin{penum}
  \item \texttt{사회복지\_관련\_법령\_법령.docx} — 환자 접근권 보장 근거 인용용
        (특히 사회복지사업법 제5조의2 등)
  \item \texttt{의료기관\_진료과목별\_전문의.xlsx} — 진료과별 전문의 수 (1,300건).
        내과(dgsbjtCd=01)·가정의학과 중심 분석
  \item \texttt{의료기관\_의료기관\_기본정보\_인력.xlsx} — 요양기관 기본정보 및
        의료인력 (10,000건). 기관 종별 분포 및 drTotCnt 활용
  \item \texttt{임상시험\_결과\_데이터.csv} — KMD-217 임상 2상 결과
        (Primary/Secondary endpoint, AE 등)
\end{penum}

\promptsec{산출물}
파일명: \texttt{KMD217\_요양급여\_등재신청서.pdf} / 8$\sim$12페이지 / 표·차트 포함 가능

\promptsec{구성 섹션}
\textbf{1) 신청 개요 (1p)}\\
신청 의약품 정보, 신청 구분(신약), 기준일. 법적 근거로 국민건강보험법 및 관련
법령 인용, 첨부 법령 문서의 사회복지사업법 제5조의2(사회복지서비스 제공의 원칙)
및 인권존중 관련 조항을 환자 접근권 보장 근거로 병기.

\textbf{2) 임상적 유용성 평가 (2$\sim$3p)}
\begin{pitemize}
  \item Primary endpoint: 24주 후 HbA1c 변화량
  \item Secondary: HbA1c $<$7.0\% 도달률, 공복혈당 변화, 체중 변화
  \item 시험군 vs.\ Sitagliptin 100mg, 95\% CI 및 p-value 포함 표 정리
  \item \textbf{비열등성 마진 $-$0.4\%(HbA1c 차이)} 기준 우월성/비열등성 판정,
        충족 항목 플래깅
  \item AE 요약
\end{pitemize}

\textbf{3) 경제성 평가 — ICER (1$\sim$2p)}
\begin{pitemize}
  \item 시험군의 QALY 개선분 대비 추가 비용으로 ICER 산출
        (비용·QALY 가정값은 명시)
  \item \textbf{WTP 임계값: 1QALY당 25,000,000원}, 미만 시 ‘비용효과적’으로 플래깅
  \item 낙관·비관 민감도 시나리오 2개 이상
\end{pitemize}

\textbf{4) 처방 환경 및 예상 시장 규모 (2$\sim$3p)}
\begin{pitemize}
  \item 진료과목별 전문의 파일에서 내과(dgsbjtCd=01)·가정의학과 전문의 수를
        기관 종별·시도별 집계
  \item 기본정보 파일에서 기관 종별 분포 및 drTotCnt 합계 산출
  \item \textbf{주요 처방 후보 기관 = 기관 종별이 ‘상급종합’ 또는 ‘종합병원’이면서
        내과 전문의 5인 이상 보유} $\rightarrow$ 시도별 표로 제시 및 플래깅
  \item 예상 처방 환자 풀 추정
\end{pitemize}

\textbf{5) 약가 산정 근거 (1p)}\\
Sitagliptin 100mg 보험약가 기준 가중평균가 산정 방식 설명(약가 가정값 명시),
임상적 우월성 반영 가산율 근거.

\medskip
\textbf{6) 결론 및 등재 권고 (0.5$\sim$1p)}

\medskip
\textbf{7) 참고문헌 / 출처}\\
첨부 파일 및 웹 보조 검색 출처 명시.

\promptsec{작성 지침}
\begin{pitemize}
  \item 한국 보건의료 등재 실무 관행 준수
  \item 수치는 한국식 단위(억 원, 만 원) 사용
\end{pitemize}
\end{taskprompt}

\noindent\begin{minipage}{\linewidth}
  \centering
  \includegraphics[width=\linewidth]{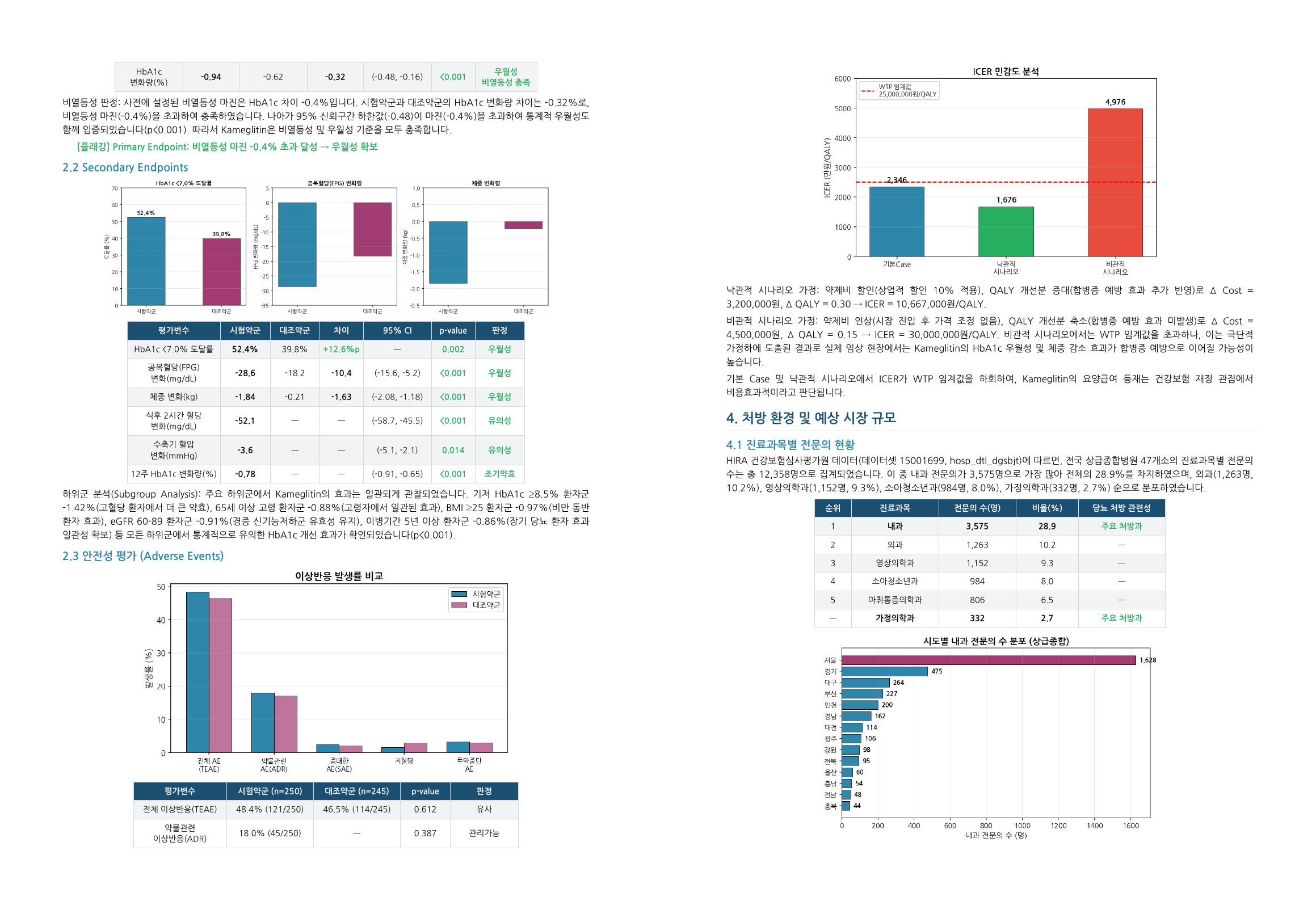}
  \captionof{figure}{Deliverable submitted by Solar Open~2:
    \texttt{KMD217\_요양급여\_등재신청서.pdf}.
  }
  \label{fig:showcase-s2}
\end{minipage}

\subsubsection{Factory energy-saving annual plan (XLSX)}

\begin{taskprompt}
당신은 경상남도 창원 공단에 위치한 중견 자동차 부품 제조사의 생산기술팀 에너지
관리 담당 과장입니다. 최근 산업용 전기요금 인상과 본사 ESG 목표(2026년
전력원단위 7\% 절감)에 따라, 1·2·3공장(사출·도장·조립 라인)의 2026년도 라인별
에너지 절감 연간계획을 한 장의 엑셀로 정리해 부문장 보고에 사용하려 합니다.

\promptsec{분석 과제}
첨부된 4개 파일을 모두 활용해 다음을 수행하세요.
\begin{penum}
  \item \textbf{현황 진단 (\texttt{에너지사용.csv} + \texttt{생산실적.csv})}
  \begin{pitemize}
    \item 공장별(1공장 INJ-A1/INJ-A2, 2공장 PNT-B1, 3공장 ASM-C1)·시간대별
          (경부하/중간부하/최대부하) 전력·LNG·스팀 사용량 합계 및 평균
          원단위(kWh당) 산출
    \item 라인별 전력원단위(kWh/개) 및 가동률(가동시간/(가동시간+정지시간)) 계산
    \item 공장별 최대부하 시간대 전력 사용 비중
  \end{pitemize}
  \item \textbf{외부 단가·시장 참조}
  \begin{pitemize}
    \item \texttt{계약종별\_전력사용량\_2023\_경상남도.xlsx} 에서 cntr=‘산업용’
          데이터의 unitCost를 powerUsage 가중평균하여 경상남도 산업용 평균 단가
          산출, \texttt{계약종별\_전력사용량\_2024\_경기도.xlsx} 의 산업용 평균
          단가와 비교
    \item \texttt{SMP\_수요예측.xlsx} 에서 areaName=‘육지’ 데이터의 hour별 평균
          SMP 곡선을 만들고, 상위 25\%를 피크, 하위 25\%를 경부하로 정의
  \end{pitemize}
  \item \textbf{2026년 절감 계획}
  \begin{pitemize}
    \item 라인별(INJ-A1, INJ-A2, PNT-B1, ASM-C1) 절감목표(\%), 핵심 투자항목,
          예상 투자비, 연간 절감금액, 단순회수기간(투자비$\div$연간절감액),
          KPI(전력원단위 kWh/개, 피크부하 비중 \%)
    \item 전사 합산 절감률 \textbf{7\% 이상} 충족
  \end{pitemize}
\end{penum}

\promptsec{산출물}
\textbf{\texttt{공장에너지절감\_2026연간계획.xlsx}} (총 2시트, 헤더 고정)
\begin{pitemize}
  \item 시트1 \texttt{현황진단}: 공장·라인·시간대별 사용량/원단위/가동률 +
        경상남도·경기도 산업용 단가 비교 + 육지 SMP 시간대별 평균
  \item 시트2 \texttt{2026절감계획}: 라인별 절감목표·투자항목·투자비·연간절감액·%
        회수기간·KPI (상단에 전사 합산 요약 행)
  \begin{pitemize}
    \item 단순회수기간 3년 이하 셀 녹색, 5년 초과 적색
    \item 금액 ₩ 천단위 콤마, 비율 소수 1자리 \%
  \end{pitemize}
\end{pitemize}

\smallskip
웹 검색을 사용한 경우 시트 하단 \texttt{참고} 영역에 출처 명시.
\end{taskprompt}

\noindent\begin{minipage}{\linewidth}
  \centering
  \includegraphics[width=\linewidth]{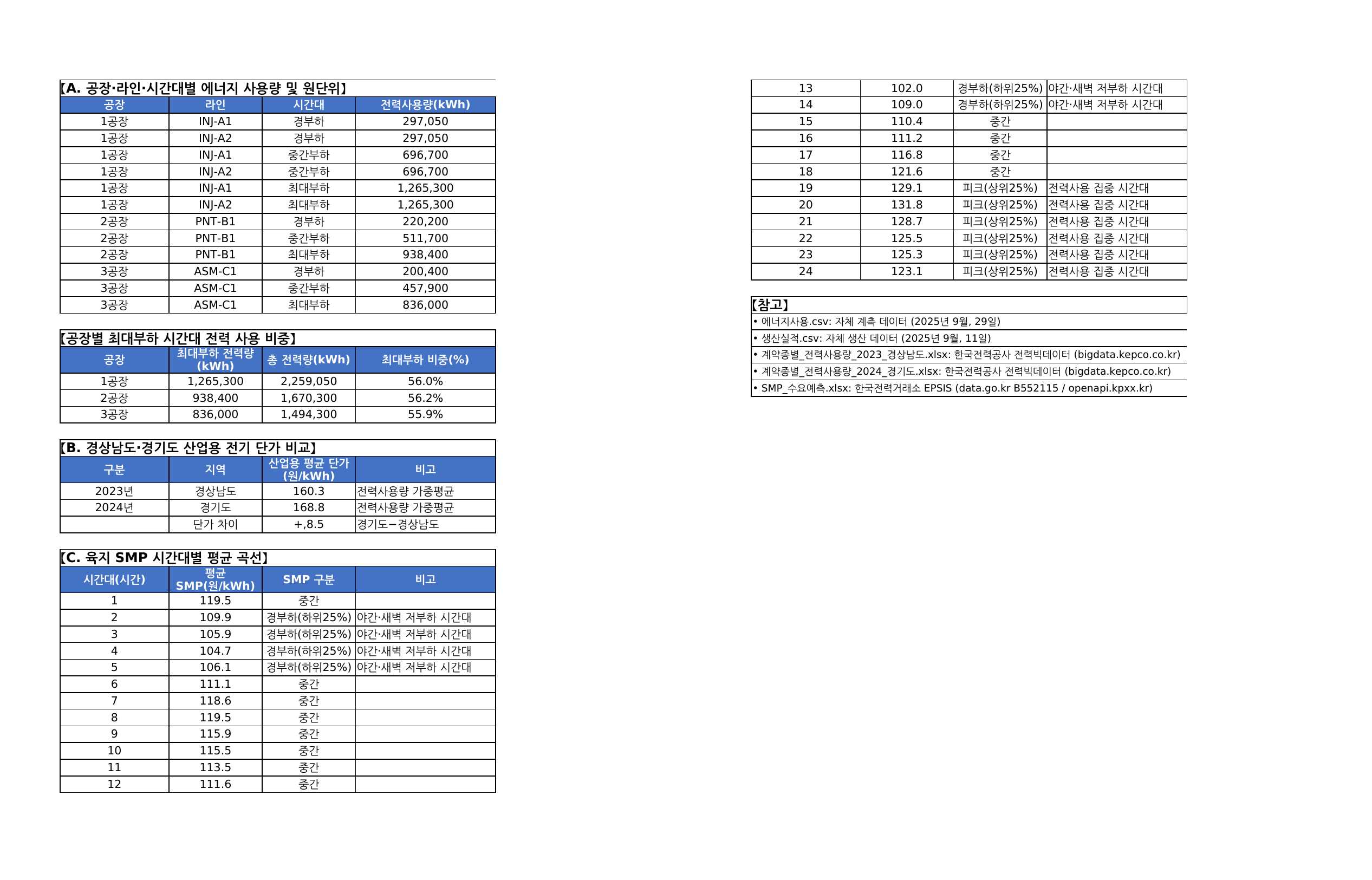}
  \captionof{figure}{Deliverable submitted by Solar Open~2:
    \texttt{공장에너지절감\_2026연간계획.xlsx}.
    }
  \label{fig:showcase-s4}
\end{minipage}

\subsubsection{Citizen briefing on national fiscal trends (DOCX)}

\begin{taskprompt}
당신은 재정 규모 약 4조 원의 광역지방자치단체 기획조정실 예산담당관실의
재정공시·주민참여예산 담당 주무관입니다. 매년 11월에 개최되는 ‘주민과 함께하는
재정설명회’를 2025년 11월 20일(목)에 앞두고, 일반 주민·시민단체·지역 언론을
청중으로 하는 브리핑 문서를 준비해야 합니다. 담당 국장은 ‘숫자 위주 결산서’가
아닌 ‘이야기가 있는 브리핑 문서’를 요구했고, 최근 10년(2015$\sim$2024년) 추이와
구조적 변화를 함께 보여달라고 지시했습니다.

첨부된 세 개의 열린재정 데이터셋(기획재정부 재정정보공개시스템 출처)을 활용해
국가 전체 재정 흐름을 우리 지자체 재정설명회의 배경 자료로 재구성하십시오.

\promptsec{첨부 파일}
\begin{penum}
  \item \texttt{재정수입구조본예산총수입기준.xlsx} — 2015$\sim$2024년 회계연도별
        본예산 총수입 구조(기금/예산, 일반회계/특별회계/기금,
        사회보험성기금·국세수입·세외수입·융자금회수·기금기타 등 세목별 금액,
        단위: 조 원)
  \item \texttt{재정지출추이총계기준.xlsx} — 2015$\sim$2024년 결산 기준 총계
        재정지출(일반회계·특별회계·기금 구분, 단위: 조 원)
  \item \texttt{조세부담률및국민부담률추이.xlsx} — 2015$\sim$2024년 결산 기준
        국세·지방세·사회보장기여금(사회보장기여금/공무원연금기여금/군인연금기여금/%
        건강보험재정기여금) 추이 및 국민부담률(\%)
\end{penum}

\promptsec{작성할 산출물}
파일명: \texttt{국가재정흐름\_주민설명\_브리핑\_2025.docx}
(한 개 문서, 분량 14$\sim$20쪽)

\promptsec{문서 구성 요구사항}
다음 6개 장(章)을 모두 포함하되, 각 장에는 평이한 서술 narrative + 표 또는
차트(이미지 또는 워드 표) + ‘주민 한마디로 정리’ 박스(2$\sim$3문장)를 함께
배치하십시오.

\medskip
\textbf{제1장. 인사말과 브리핑의 목적 (1쪽)}
\begin{pitemize}
  \item 재정설명회의 취지, 본 문서가 다루는 기간(2015$\sim$2024년)과 데이터
        출처(열린재정 시스템)를 명시
\end{pitemize}

\textbf{제2장. 나라 살림의 ‘수입’ 이야기 (3$\sim$4쪽)}
\begin{pitemize}
  \item \texttt{재정수입구조본예산총수입기준.xlsx} 활용
  \item 2015년과 2024년의 총수입 규모(조 원)를 비교하고, 10년간 증감액·증감률 산출
  \item 회계구분(BDG\_FND\_DIV\_NM: 기금/예산)과 세부구분(ACNT\_DIV\_NM:
        일반회계/기업특별회계/기타특별회계/기금)별 비중 변화 표로 정리
  \item 세목(SMOK\_DIV\_NM)별 2024년 기준 상위 5개 항목을 막대그래프로 시각화
  \item ‘사회보험성기금’의 절대 규모가 10년간 어떻게 변했는지 별도 단락으로 강조
\end{pitemize}

\textbf{제3장. 나라 살림의 ‘지출’ 이야기 (3$\sim$4쪽)}
\begin{pitemize}
  \item \texttt{재정지출추이총계기준.xlsx} 활용
  \item 2015$\sim$2024년 일반회계·특별회계·기금 결산지출 추이 선그래프(3개 시리즈)
  \item 연도별 총지출 합계 표 + 전년대비 증감률 컬럼
  \item 코로나 시기(2020$\sim$2022년) 지출 급증 구간을 narrative로 설명하고,
        2023년 일반회계 지출이 전년 대비 어떻게 변했는지 별도 코멘트(증감액·증감률 명기)
  \item 10년간 일반회계 대비 기금 지출 비율이 어떻게 바뀌었는지 분석
\end{pitemize}

\textbf{제4장. ‘내가 낸 세금’은 어디로 가나 — 부담률 이야기 (3$\sim$4쪽)}
\begin{pitemize}
  \item \texttt{조세부담률및국민부담률추이.xlsx} 활용
  \item 국세·지방세·사회보장기여금의 10년 추이 표 및 누적영역차트
  \item 사회보장기여금 내 4개 세부항목(사회보장기여금/공무원연금기여금/%
        군인연금기여금/건강보험재정기여금) 중 건강보험재정기여금의 증가 속도를
        narrative로 설명
  \item 2024년 국민부담률(\%) 수치를 인용하고, 이것이 주민 개인 입장에서 어떤
        의미인지 평이한 비유(예: ‘소득 100만 원 중 OO원’)로 풀어쓰기
  \item 2020년부터 데이터에 등장하는 ‘국세’ 항목과 그 이전 시기의 데이터 가용성
        차이를 각주로 안내
\end{pitemize}

\textbf{제5장. 우리 지자체 재정설명회에 시사하는 점 (2$\sim$3쪽)}
\begin{pitemize}
  \item 위 세 데이터에서 도출한 국가 재정의 3대 구조적 변화(① 사회보험성기금
        비중 확대, ② 기금 지출의 지속적 증가, ③ 국민부담률 상승 추세)를 정리
  \item 각 변화가 광역지자체(재정규모 약 4조 원) 입장에서 어떤 함의를 갖는지
        1$\sim$2문단씩 서술(지방세 의존도, 사회복지 매칭부담, 국고보조금 의존 등
        관점에서)
  \item 단, 우리 지자체의 구체 결산 수치는 본 문서에 등장하지 않으며,
        ‘국가 재정 맥락’만 다룬다는 점을 본문에 명시
\end{pitemize}

\textbf{제6장. 부록: 데이터 출처 및 용어 풀이 (1$\sim$2쪽)}
\begin{pitemize}
  \item 사용 데이터셋 3건의 출처(열린재정, 기획재정부 재정정보공개시스템) 표기
  \item 일반 주민이 헷갈리기 쉬운 용어 8개 이상을 ‘한 줄 풀이’로 정리(예:
        일반회계, 특별회계, 기금, 본예산, 결산, 총계기준, 국민부담률, 조세부담률,
        사회보장기여금 등)
  \item 웹에서 보조적으로 용어 정의나 통계 관행을 참조한 경우 마지막에
        ‘Sources’ 절을 두고 URL과 함께 인용
\end{pitemize}

\promptsec{수치 검증 원칙}
\begin{pitemize}
  \item 모든 금액 단위는 ‘조 원’(첨부파일 단위 유지), 비율은 소수점 첫째
        자리까지 표기
  \item 증감률은 ((당해연도 $-$ 기준연도) / 기준연도) $\times$ 100 공식으로
        산출하고 소수점 첫째 자리까지
  \item 차트는 워드 내장 표 또는 이미지 캡션과 함께 삽입, 각 차트에는
        ‘그림 N. 제목 (출처: 파일명)’ 형식의 캡션 부착
  \item 본문 내 인용한 모든 수치는 첨부 파일에서 확인 가능한 값과 일치해야 함
\end{pitemize}

\promptsec{톤앤매너}
\begin{pitemize}
  \item 회계 비전공 주민 청중을 위해 평이한 구어체와 비유 사용
        (단, 결산·예산·기금 등 기본 용어는 정확히)
  \item 정치적 평가나 정책 옹호 표현은 배제하고, 사실 중심 서술 유지
  \item 각 장 말미의 ‘주민 한마디로 정리’ 박스는 회색 음영 또는 박스 테두리로 구분
\end{pitemize}

\promptsec{추가 요구사항}
\begin{pitemize}
  \item 모든 차트·표에는 ‘그림 N. 제목 (출처: 파일명)’ 형식의 캡션을 부착하고,
        본문에서 그림 번호로 인용할 것
  \item 제2$\sim$5장 말미에 회색 음영 또는 박스 테두리로 구분된
        ‘주민 한마디로 정리’ 박스(2$\sim$3문장)를 배치
  \item 제6장에 일반 주민용 용어 풀이를 최소 8개 이상 한 줄 정의 형식으로 수록
  \item 웹 보조 참조를 사용한 경우에 한해 문서 말미 ‘Sources’ 절에 URL과
        접근일자를 함께 명기
  \item 산출물 규모 제한: 14쪽 이상 20쪽 이하.
  \item 기간 제약: 2015$\sim$2024년 회계연도 10년치 데이터 전체를 반드시 다룰 것.
  \item 기간 제약: 2025년 11월 20일(목) 재정설명회 전일까지 제출.
  \item 준수 기준: 열린재정 데이터 표기 관행. 금액 단위는 ‘조 원’(원 데이터 단위
        유지), 비율은 \% 단위, 모두 소수점 첫째 자리까지 표기
\end{pitemize}
\end{taskprompt}

\noindent\begin{minipage}{\linewidth}
  \centering
  \includegraphics[width=\linewidth]{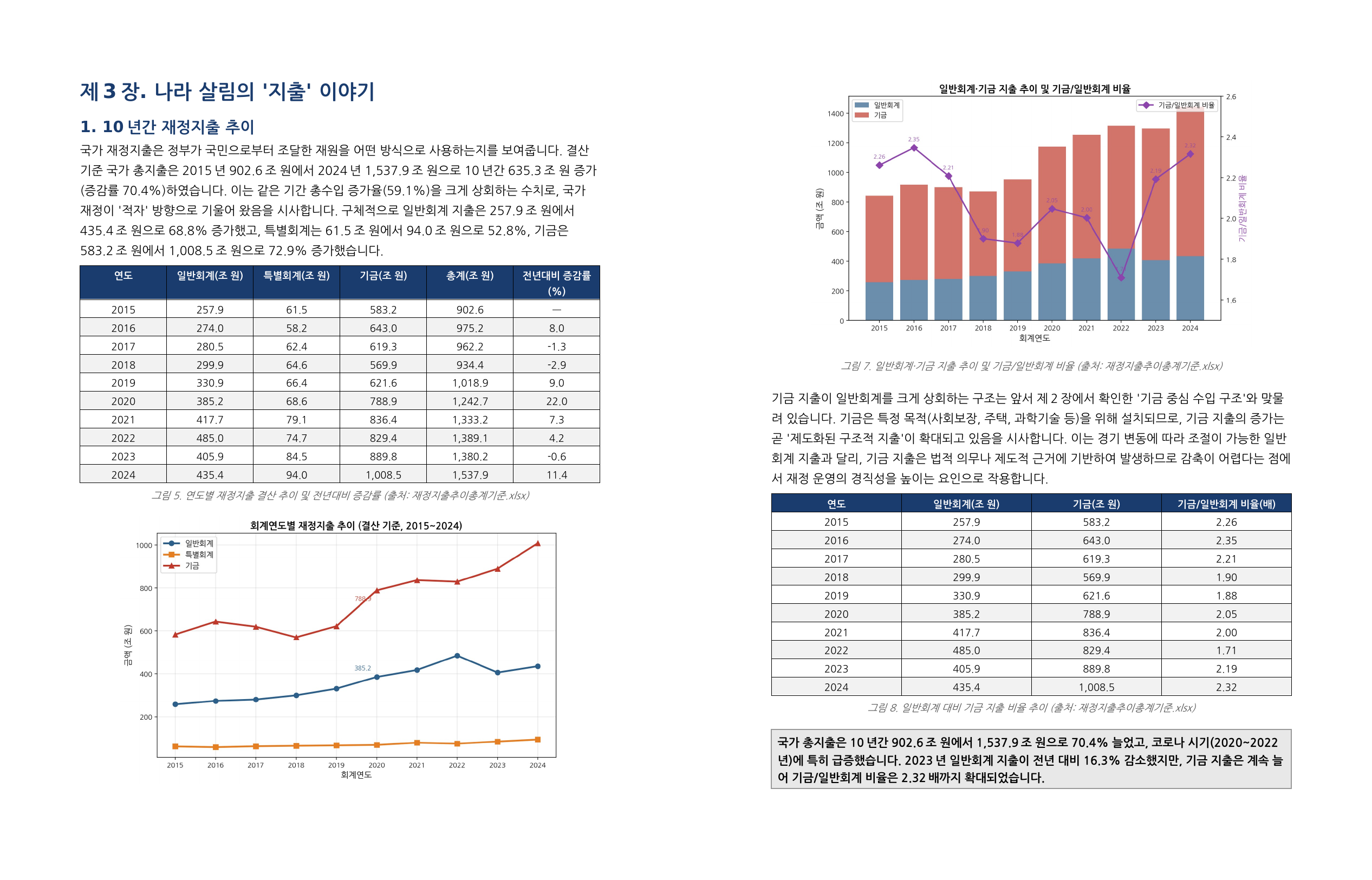}
  \captionof{figure}{Deliverable submitted by Solar Open~2:
    \texttt{국가재정흐름\_주민설명\_브리핑\_2025.docx}.
    }
  \label{fig:showcase-s5}
\end{minipage}

\subsubsection{Municipal budget-committee response deck (PPTX)}

\begin{taskprompt}
당신은 한솔시 기획예산담당관실의 예산팀 주무관(예산 분석 담당)입니다. 다음 주
개최되는 한솔시의회 예산결산특별위원회 정례 심사에서, 우리 시의 2024 회계연도
예산 집행 현황을 보고하고 의원 질의에 대비한 부서 답변 자료를 발표해야 합니다.
예결특위 위원들은 단순 집행률 수치를 넘어 ‘중앙정부 재정 흐름과 연계한 지방재정
운용의 합리성’을 추궁할 것으로 예상되므로, 국가 재정지출·국세수입 추이를 거시
배경으로 활용하여 우리 시 사업별 집행 실적과 미집행 사유를 설득력 있게 정리해야
합니다.

\promptsec{활용 자료}
\begin{pitemize}
  \item \texttt{budget\_line\_items.xlsx} — 우리 시 2024 회계연도 사업·세목별
        예산액·집행액·잔액·집행률 (요약/세목 명세 2개 시트)
  \item \texttt{재정지출추이본예산총지출기준.xlsx} — 열린재정 국가 재정지출 추이
        (2015$\sim$2024, 일반회계·특별회계·기금)
  \item \texttt{국세수입추이.xlsx} — 열린재정 국세수입 추이 (2015$\sim$2024, 세목별)
  \item \texttt{재정수입구조본예산총수입기준.xlsx} — 열린재정 국가 재정수입 구조
        (2015$\sim$2024)
  \item \texttt{dept\_profile.md} — 한솔시 기획예산담당관실 기관 프로필
\end{pitemize}

\promptsec{산출물}
\texttt{한솔시\_2024회계연도\_예결특위\_답변자료.pptx} — 10$\sim$13장 분량의
발표용 슬라이드. 모든 슬라이드에는 의원 질의 대응에 활용할 수 있는 상세 스피커
노트(slide notes)를 포함하십시오.

\promptsec{필수 구성 (슬라이드 구조)}
\begin{penum}
  \item \textbf{표지} — 한솔시 기획예산담당관실, 2024 회계연도 예결특위
        답변자료, 발표일자(2025년 6월 17일 기준)
  \item \textbf{거시 재정 배경 (1)} — 국가 재정지출 추이 (2015$\sim$2024,
        일반회계·특별회계·기금 별 추이 그래프).
        \texttt{재정지출추이본예산총지출기준.xlsx} 기반.
  \item \textbf{거시 재정 배경 (2)} — 국세수입 추이 및 주요 세목 비중 변화
        (2020$\sim$2024 집중). \texttt{국세수입추이.xlsx} 기반. 지방교부세 재원인
        내국세 흐름을 강조.
  \item \textbf{거시 재정 배경 (3)} — 국가 재정수입 구조 변화의 시사점
        (사회보험성기금·세외수입·국세수입 추이).
        \texttt{재정수입구조본예산총수입기준.xlsx} 기반.
  \item \textbf{한솔시 2024 예산 총괄} — 분야별 예산액·집행액·집행률 (10개 분야
        막대그래프). \texttt{budget\_line\_items.xlsx} 요약 시트 기반. 시 전체
        평균 집행률과 함께 표기.
  \item \textbf{집행률 우수 분야 / 부진 분야} — 집행률 75\% 이상 분야와 65\%
        미만 분야를 색상 구분(우수: 녹색, 부진: 적색)하여 비교.
  \item \textbf{사회복지비 세부 분석} — SOC-2024-001 산하 세목별 집행 현황.
        세목 명세 시트 기반.
  \item \textbf{지역경제·일자리 분야 심층 분석} — ECN-2024-001, ECN-2024-002
        세목별 부진 사유 분석.
  \item \textbf{SOC인프라(도로·교통, 상하수도) 미집행 사유} — INF-2024-001,
        INF-2024-002 분석.
  \item \textbf{거시 재정과 지방 집행 연계 진단} — 국세수입 둔화 $\rightarrow$
        지방교부세 교부 지연 $\rightarrow$ 시 사업 집행 부진의 연결 논리를
        1장으로 정리.
  \item \textbf{향후 집행 계획 및 이월 최소화 대책} — 분야별 4분기 집행 가속화 방안.
  \item \textbf{예상 질의응답(Q\&A)} — 의원 예상 질의 5개와 답변 요지.
\end{penum}

\promptsec{분석 요건}
\begin{pitemize}
  \item 집행률은 \texttt{집행액 / 예산액}으로 계산하되, 원본 시트의
        \texttt{집행률} 열과 교차검증하십시오. 소수점 첫째 자리(\%)까지 표시.
  \item 분야별 잔액(미집행액)을 억 원 단위로 환산하여 표기
        (예: 6,555,000,000원 $\rightarrow$ 65.55억 원).
  \item 거시 재정 추이 그래프는 2015$\sim$2024 전체 10개년을 표시하되, 시사점
        도출은 최근 5개년(2020$\sim$2024)에 집중.
  \item 국세수입 분석 시 ‘내국세’ 관련 세목(소득세·법인세·부가가치세 등)을 우선
        강조 (지방교부세 재원이기 때문).
\end{pitemize}

\promptsec{제약 조건}
\begin{pitemize}
  \item 집행률 70\% 미만인 분야는 별도 표시(적색)하고, 미집행 사유를 본문에
        1줄 이상 명시할 것.
  \item 모든 금액 단위는 슬라이드 본문에서 ‘억 원’으로 통일(원 단위 그대로 노출
        금지). 단, 부록 표에서는 원 단위 병기 허용.
  \item 슬라이드는 총 10$\sim$13장 범위. 표지·목차 제외하고 본문은 시각
        자료(그래프·표)를 최소 6장 이상 포함.
  \item 모든 슬라이드에 스피커 노트를 작성하되, 노트에는 의원이 던질 만한 추궁
        포인트와 대응 논리를 포함할 것.
  \item 외부 자료(법령·통계 정의 등)를 보조적으로 인용한 경우 마지막 슬라이드에
        ‘Sources’ 섹션을 두어 출처를 기재할 것. 첨부 파일 자체는 별도 출처 표기
        불필요.
\end{pitemize}

\smallskip
발표 대상일은 2025년 6월 17일이며, 회계연도는 2024년 1월 1일 $\sim$ 2024년
12월 31일 기준입니다.

\promptsec{추가 요구사항}
\begin{pitemize}
  \item 각 슬라이드에 의원 추궁 포인트와 대응 논리를 담은 스피커 노트 작성
  \item 집행률 75\% 이상 분야는 녹색, 70\% 미만 분야는 적색으로 시각화
  \item 슬라이드 본문 금액은 모두 억 원 단위로 표기 (소수점 둘째 자리까지)
  \item 산출물 규모 제한: 표지 포함 10$\sim$13장.
  \item 사용할 방식: 집행률 교차검증 (집행액/예산액으로 직접 계산한 값과 원본
        시트의 집행률 열을 비교하여 일치 여부 확인).
  \item 기간 제약: 한솔시 회계연도 2024년 1월 1일 $\sim$ 2024년 12월 31일.
  \item 기간 제약: 2025년 6월 17일 예결특위 정례 심사.
  \item 기간 제약: 거시 재정 추이는 2015$\sim$2024 전체, 시사점은
        2020$\sim$2024에 집중.
  \item 플래깅 기준: 집행 부진 분야 표시 기준 less\_than 집행률 70\%인 항목을
        표시하십시오.
  \item 플래깅 기준: 집행 우수 분야 표시 기준 greater\_than\_or\_equal 집행률
        75\%인 항목을 표시하십시오.
  \item 다음 항목을 포함하십시오: 사회복지비(SOC-2024-001),
        보건의료비(SOC-2024-002), 지역경제활성화(ECN-2024-001),
        일자리창출(ECN-2024-002), 도로및교통(INF-2024-001),
        상하수도(INF-2024-002), 환경보전(ENV-2024-001), 교육지원(EDU-2024-001),
        문화체육관광(CUL-2024-001), 일반행정운영(GEN-2024-001).
  \item 순서 제약: 거시 재정 배경 슬라이드(2$\sim$4장) $\rightarrow$ 한솔시 예산
        분석 슬라이드(5장 이후).
  \item 순서 제약: 분야별 집행 현황 진단 $\rightarrow$ 거시-지방 연계 진단 및
        향후 계획.
  \item 사회복지비 세부 분석 슬라이드에는 SOC-2024-001 산하 5개 세목을 모두 포함
\end{pitemize}
\end{taskprompt}

\noindent\begin{minipage}{\linewidth}
  \centering
  \includegraphics[width=\linewidth]{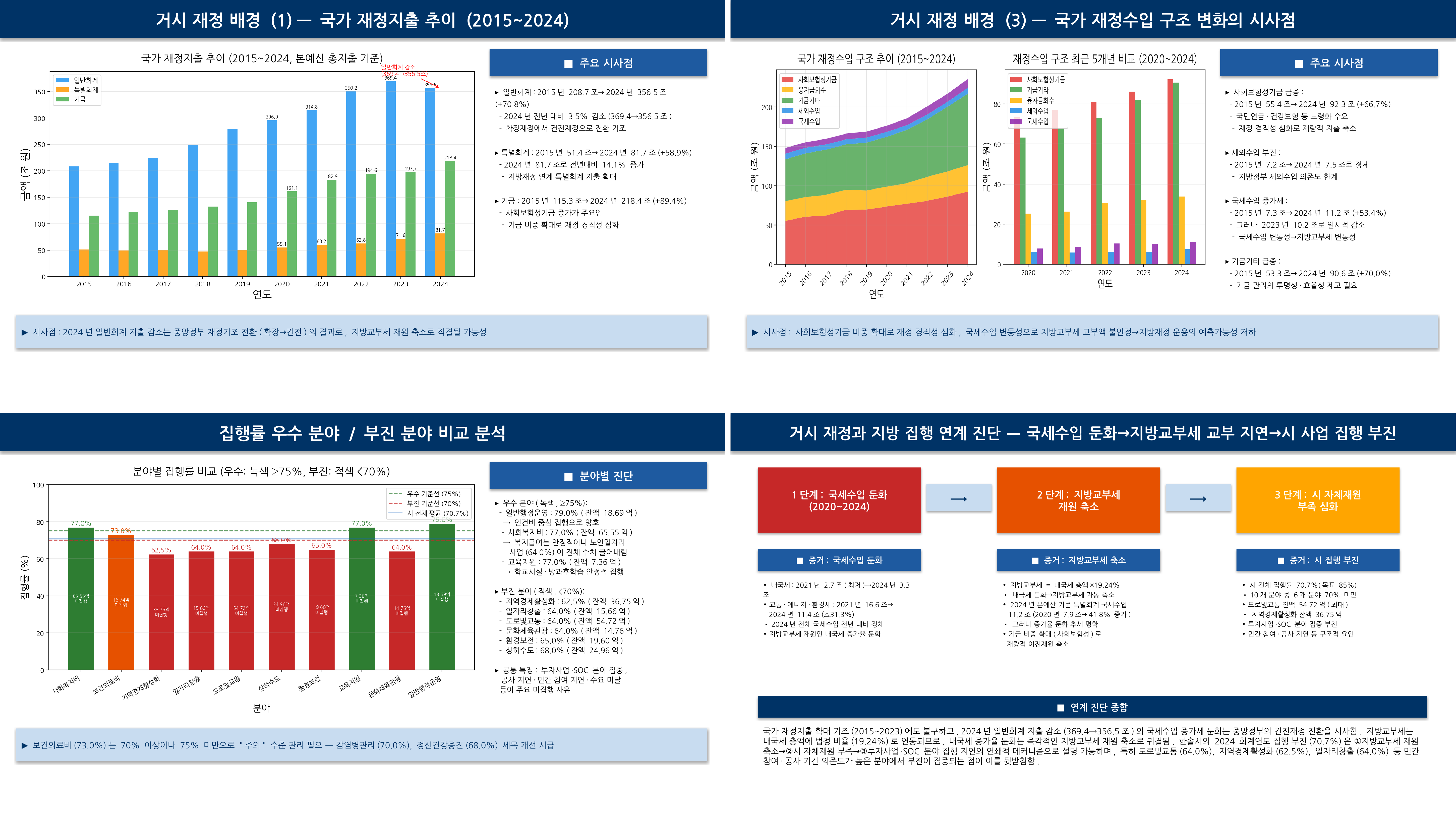}
  \captionof{figure}{Deliverable submitted by Solar Open~2:
    \texttt{한솔시\_2024회계연도\_예결특위\_답변자료.pptx}.
    }
  \label{fig:showcase-s7}
\end{minipage}

\end{document}